%% file: main-iclr.tex
\PassOptionsToPackage{table,xcdraw}{xcolor}

\documentclass{article} 
\usepackage{iclr2026_conference,times}

\usepackage[utf8]{inputenc} 
\usepackage[T1]{fontenc}    
\usepackage{hyperref}       
\usepackage{url}            
\usepackage{booktabs}       
\usepackage{amsfonts}       
\usepackage{nicefrac}       
\usepackage{microtype}      
\usepackage{xspace}
\usepackage{amsmath}
\usepackage{graphicx}
\usepackage{colortbl}
\usepackage{multirow}
\usepackage[normalem]{ulem}
\useunder{\uline}{\ul}{}
\usepackage{arydshln}
\usepackage{comment}
\usepackage{makecell}

\newcommand{\deltapos}[1]{{\scriptsize\textcolor{blue}{$\uparrow$#1}}}
\newcommand{\deltaneg}[1]{{\scriptsize\textcolor{red}{$\downarrow$#1}}}

\newcommand{\sys}{{GUI-Shift}\xspace}

\title{%
  \sys: Enhancing VLM-Based GUI Agents through Self-supervised Reinforcement Learning%
}

\author{
    \textbf{Longxi Gao$^{1}$, Li Zhang$^{1}$, Pengzhi Gao$^{2}$, Wei Liu$^{2}$, Jian Luan$^{2}$, Mengwei Xu$^{1}$} \\
    $^1$\textmd{Beijing University of Posts and Telecommunications} 
    $^2$\textmd{Unaffiliated} \\
    \texttt{glx@bupt.edu.cn}
}

\iclrfinalcopy 

\begin{document}

\maketitle

\input{sections/abstract.tex}
\input{sections/intro.tex}
\input{sections/related_work.tex}
\input{sections/method.tex}
\input{sections/experiments.tex}
\input{sections/ablation.tex}
\input{sections/conclusion.tex}

\newpage
\bibliographystyle{iclr2026_conference}
\bibliography{reference}

\appendix
\newpage
\input{sections/appendix.tex}

\end{document}

%% file: sections/abstract.tex
\begin{abstract}
    Training effective Vision-Language Models (VLMs) for GUI agents typically depends on large-scale annotated datasets, whose collection is both labor-intensive and error-prone.
    We introduce \textbf{\(K\)-step GUI Transition}, a self-supervised inverse dynamics task in which VLMs learn GUI dynamics by predicting the initial action that causes a transition between two GUI states.
    This approach eliminates the need for natural language instructions and enables scalable dataset construction from existing GUI trajectories or automated exploration.
    Building on this task, we propose \textbf{\sys}, a reinforcement learning (RL) framework that combines rule-based optimization with data filtering to improve VLM performance.
    We conduct extensive experiments using multiple VLM backbones across four benchmarks, spanning GUI task automation (AndroidControl, GUI Odyssey) and GUI grounding (ScreenSpot-v2, ScreenSpot-Pro).
    Our results show that training on \sys generalizes well to both GUI automation and grounding tasks, yielding up to an 11.2\% increase in GUI automation accuracy.
    This study underscores the potential of self-supervised RL to leverage unlabeled GUI trajectories and offers a scalable alternative to training with annotated samples.
  \end{abstract}

%% file: sections/intro.tex
\section{Introduction}

Mobile GUI agents~\citep{gou2025navigatingdigitalworldhumans,hong2024cogagentvisuallanguagemodel,qin2025uitarspioneeringautomatedgui,wen2024autodroidllmpoweredtaskautomation,yang2024ariauivisualgroundinggui} 
interpret natural language instructions and perform actions (e.g., click, scroll) directly on smartphone screens.
They can control diverse apps as a human would, improving accessibility for users who are visually impaired, elderly, or have their hands occupied.
Breakthroughs of vision language models (VLMs)~\citep{bai2025qwen25vltechnicalreport,chen2024internvlscalingvisionfoundation,coreteam2025mimovltechnicalreport} have reshaped the design paradigm of mobile GUI agents, transitioning from handcrafted heuristics to learned, vision-grounded policies.
However, VLMs still struggle to deliver satisfactory accuracy~\citep{dai2025advancingmobileguiagents,qin2025uitarspioneeringautomatedgui,rawles2024androidworld,zhang2025doeschainofthoughtreasoninghelp}, especially when facing complex multi-step tasks.
A common approach for enhancing VLMs is through supervised fine-tuning (SFT) on datasets containing GUI interaction trajectories paired with human-annotated task instructions~\citep{li2024effectsdatascaleui,rawles2023androidwildlargescaledataset}.
Yet effective, collecting GUI trajectories with task instructions remains labor-intensive and error-prone~\citep{Deka:2017:Rico,rawles2023androidwildlargescaledataset}.
For example, the AndroidControl~\citep{li2024effectsdatascaleui} dataset takes one year of paid annotation effort to produce just 15,283 task demonstrations.
Such high annotation cost limits the scalability of this paradigm.

In this study, we aim to address a fundamental challenge: \textit{how to train capable mobile GUI agents using large-scale, unlabeled GUI trajectories, rather than relying on costly human-annotated instructions.}
To tackle this, we propose a self-supervised training task, termed \textit{\(K\)-step GUI Transition}.
Inspired by inverse-dynamics modeling in robotics and biomechanics~\citep{brandfonbrener2023inverse, tian2024predictive, zapolsky2016inversedynamicsrigidcontact}, where a model predicts control commands linking two consecutive physical states, our task treats screenshots as states and GUI actions as commands. 
Each training sample in \(K\)-step GUI Transition consists of two screenshots, \(S_t\) and \(S_{t+k}\), where \(S_{t+k}\) results from executing \(k\) actions starting from \(S_t\).
The VLM is trained to predict the first action that transforms \(S_t\) into \(S_{t+1}\). 
This design offers two key advantages:
(1) \textit{Explicit state-change signal.} Each sample contains a pair of GUI screenshots, enabling the model to utilize inter-screen visual differences and temporal cues, rather than learning from a single screen.
(2) \textit{Efficient data utilization at scale.} 
Since ground-truth actions are embedded in GUI trajectories~\citep{rawles2023androidwildlargescaledataset,li2024effectsdatascaleui}, no predefined instructions or manual annotations are needed.
Moreover, for any \(k\), a GUI trajectory with \(n\) screens can yield up to \(n-k\) training samples, enabling scalable data construction.
These benefits make \(K\)-step GUI Transition a strong candidate for self-supervised GUI agent training.

With the self-supervised training task, it is essential to determine how to effectively enhance VLMs.
In GUI tasks, multiple action parameters can often be functionally equivalent and result in the same next state.
For example, any coordinate within a button's bounding box is valid for a click, and textual inputs may be accepted in various formats or with different keywords.
This multiplicity makes supervised fine-tuning (SFT) suboptimal, as it enforces a single reference action in a static dataset through the cross-entropy loss, penalizing all other valid alternatives and therefore providing misleading learning signals. 
To address this limitation, we adopt Group Relative Policy Optimization (GRPO)~\citep{shao2024deepseekmathpushinglimitsmathematical}, which samples diverse plausible actions, evaluates them using a task-specific reward function, and ranks them based on group-normalized advantages.
For example, in click actions, rewards are assigned if the sampled point lies within the target bounding box, offering a more tolerant and informative optimization signal.
Overall, GRPO provides a more effective training paradigm for GUI agents by encouraging exploration and increasing robustness to action variability.

To this end, we present \sys, a self-supervised reinforcement learning (RL) framework that applies GRPO to \(K\)-step GUI Transition.
Figure~\ref{fig:overview} illustrates the overview of the \sys framework.
To select data matched to the model's learning ability, we adopt a unified action sampling and scoring mechanism during both data filtering and training stages.
For each sample, the VLM generates \(N\) action predictions, each scored based on format and action correctness.
Only samples containing both correct and incorrect predictions among the \(N\) predictions are selected for training.
After training, the VLM acquires GUI-specific capabilities and serves as a more effective backbone for GUI agents.
VLMs enhanced with \sys also have the ability to generalize well to GUI task automation and GUI grounding tasks without further alignment or fine-tuning.

\input{figs/overview.tex}
We apply \sys to train four VLMs: Qwen2.5-VL-7B~\citep{bai2025qwen25vltechnicalreport}, InternVL3-8B~\citep{chen2024internvlscalingvisionfoundation}, MimoVL-7B-SFT~\citep{coreteam2025mimovltechnicalreport}, and MimoVL-7B-RL, each using 2K samples for four \(K\)-step GUI Transition variants (\(k \in \{1,2,3,4\}\)).
We evaluate the VLMs on four benchmarks: AndroidControl~\citep{li2024effectsdatascaleui} and GUI Odyssey~\citep{lu2024guiodyssey} for GUI task automation, and ScreenSpot-v2~\citep{wu2024osatlasfoundationactionmodel} and ScreenSpot-Pro~\citep{li2025screenspotproguigroundingprofessional} for GUI grounding.
Overall, VLMs enhanced with \sys show notable improvements over their base versions.
For example, \sys-Qwen achieves up to 11.2\% higher accuracy on AndroidControl-High and 2.5\% on ScreenSpot-v2, yielding 70.4\% and 89.0\% overall accuracy on the respective benchmarks.
We also conduct comprehensive ablation studies to examine the effects of data filtering, task formulation, reasoning configurations, and training paradigms.
The results show that using the target state \(S_{t+k}\) as a visual instruction offers an effective alternative to human- or AI-annotated textual instructions for training GUI agents.
The key contributions are summarized below.

\begin{enumerate}
    \renewcommand{\labelenumi}{(\arabic{enumi})}
    \item We introduce \(K\)-step GUI Transition, a training task that leverages abundant unlabeled GUI trajectories to enhance VLMs used in GUI agents.
    \item We propose \sys, a self-supervised RL framework that bridges the gap between GUI dynamics modeling and action-level GUI learning, mitigating the limitation of SFT in handling action multiplicity and poor generalization in GUI tasks.
    \item Experiments across four VLMs and four benchmarks show that VLMs enhanced with \sys exhibit generalization in both GUI automation and grounding tasks, with up to 11.2\% accuracy gains.
  \end{enumerate}

%% file: figs/overview.tex
\begin{figure}[t]
    \centering
    \includegraphics[width=\linewidth]{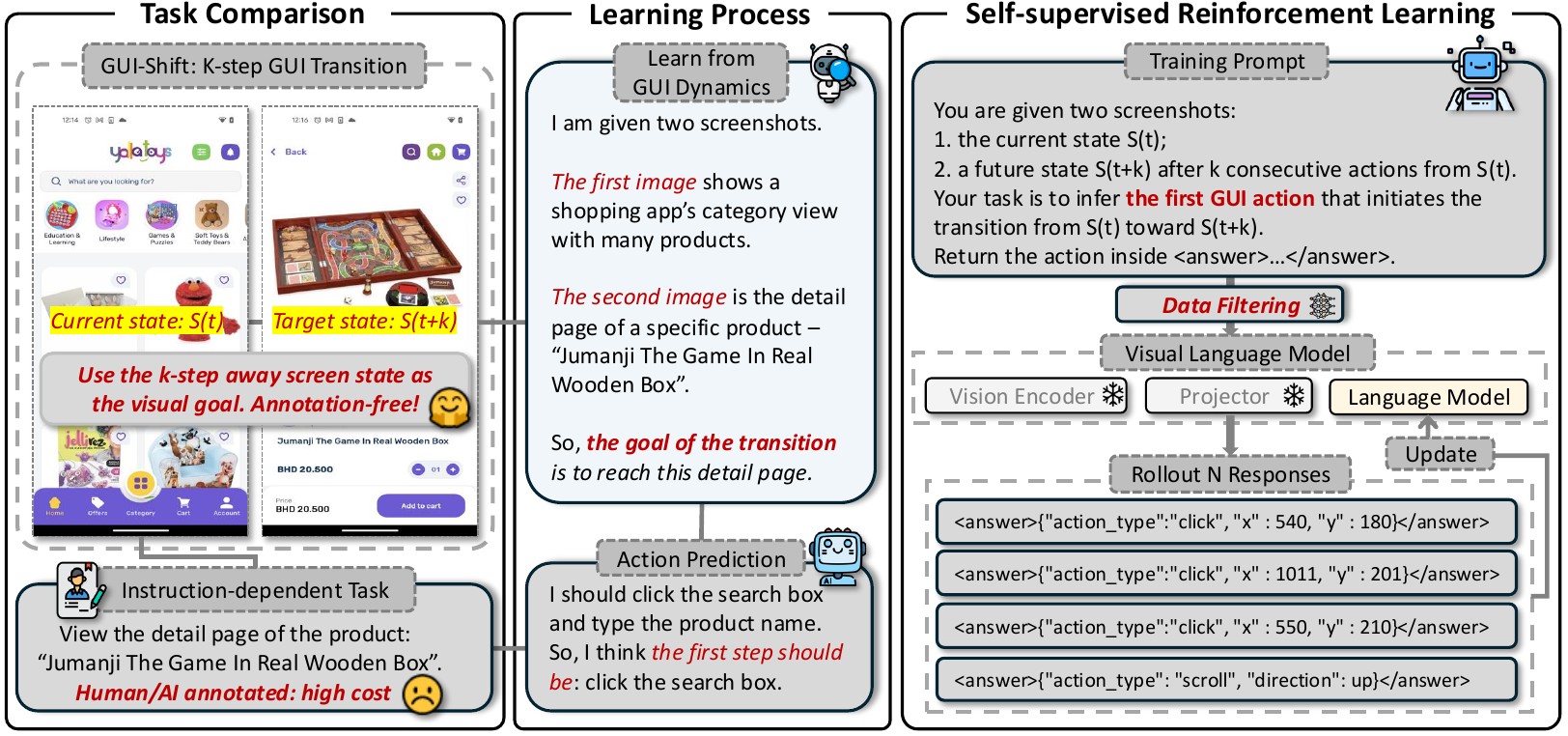}
    \caption{
      Overview of the \sys framework.
      \textbf{Left}: \(K\)-step GUI Transition replaces annotated instructions with the target state \(S_{t+k}\), enabling scalable data construction through automated offline exploration.
      \textbf{Middle}: The model learns GUI dynamics by predicting the action that causes the transition.
      \textbf{Right}: \sys achieves self-supervised training by applying GRPO to GUI Transition.
    }
    \label{fig:overview}
  \end{figure}
  

%% file: sections/related_work.tex
\section{Related Work}

\subsection{Mobile GUI Agents}
Recent progress in mobile GUI agents has been driven by VLMs trained via SFT on large-scale datasets.
These models learn to map instructions to GUI actions using instruction-following tasks, making 
high-quality annotations essential.
Despite the availability of diverse GUI datasets~\citep{Deka:2017:Rico,gao2024mobileviewslargescalemobilegui,li2024effectsdatascaleui,lu2024guiodyssey,rawles2023androidwildlargescaledataset}, 
the quantity of high-quality annotations remains insufficient for robust training and usually requires significant human effort to scale.
To reduce annotation costs, prior pipelines often incorporate out-of-domain image-caption pairs~\citep{hong2024cogagentvisuallanguagemodel,wang2023cogvlm}
and supplement training with web and desktop data to improve cross-platform generalization~\citep{cheng2024seeclickharnessingguigrounding}.
As a result, the overall scale of training data tends to be large:
Uground~\citep{gou2025navigatingdigitalworldhumans} uses 1.3M screenshots to train a visual grounding model; OS-Atlas~\citep{wu2024osatlasfoundationactionmodel} leverages 13M GUI elements for grounding pretraining. 
Some recent approaches have explored GUI state modeling.
UI-TARS~\citep{qin2025uitarspioneeringautomatedgui} incorporates a state transition task, which focuses on describing visual changes between screenshots rather than predicting the underlying actions, resulting in a gap with GUI task automation.  
MobileVLM~\citep{wu2024mobilevlmvisionlanguagemodelbetter} introduces an action prediction task between screenshots, but is restricted to one-step transitions and SFT.
They still rely on annotation-heavy fine-tuning for downstream alignment and generalization.
In this work, we propose \(K\)-step GUI Transition, formulating a \(k\)-step inverse dynamics objective that enables scalable training on large, unlabeled, and underutilized GUI datasets.

\subsection{Rule-based Reinforcement Learning}
Rule-based RL has proven to be a promising alternative to SFT.
GRPO~\citep{shao2024deepseekmathpushinglimitsmathematical} uses a reward model to score each response and computes group relative advantages
instead of training a critic model, whose size is comparable to the policy model,
thereby significantly reducing computational cost.
Reinforcement Learning with Verifiable Rewards ~\citep{lambert2025tulu3pushingfrontiers} further emphasize the use of verifiable answers to design reliable reward signals.
DeepSeek-R1~\citep{deepseekai2025deepseekr1incentivizingreasoningcapability} shows that simple format and accuracy rewards 
are sufficient to surpass the performance of instruction-tuned models.
Several recent works have applied GRPO on GUI tasks:
UI-R1~\citep{lu2025uir1enhancingactionprediction} employs one-stage RL on 136 samples with step-level instructions.
GUI-R1~\citep{luo2025guir1generalistr1style} expands this to 3K task-level instructions from five platforms.
InfiGUI-R1~\citep{liu2025infiguir1advancingmultimodalgui} adopts a two stage SFT+RL pipeline and scales to 32K samples from both GUI and non-GUI domains.
UI-Venus~\citep{gu2025uivenus} employs GRPO to two variants, using 107K samples for grounding and 350K for navigation.
While these works demonstrate the effectiveness of rule-based RL for GUI agents,
they still rely on annotated instructions and require reasoning during training and inference.
Different from these annotation-dependent training paradigms, \sys fine-tunes VLMs via one-stage RL on \(K\)-step GUI Transition in a self-supervised manner,
achieving competitive performance and demonstrating strong generalization across GUI task automation and GUI grounding benchmarks.

%% file: sections/method.tex
\section{Methodology}
\label{sec:framework}
\sys is a self-supervised RL framework designed to enhance VLM-based GUI agents through the \(K\)-step GUI Transition task.
In this section, we first describe GRPO, the underlying training algorithm in \sys.
We then detail the reward design tailored to GUI action modeling,
and present the complete \sys framework along with its rationale and advantages.

\subsection{Preliminaries}
\label{sec:preliminaries}
GRPO~\citep{shao2024deepseekmathpushinglimitsmathematical} offers a computationally efficient alternative to Proximal Policy Optimization (PPO)~\citep{schulman2017proximalpolicyoptimizationalgorithms}, a widely used actor-critic method.
Instead of maintaining a separate critic network for value estimation,
GRPO computes normalized, group-wise advantages \(A_i\) directly from reward scores,
thereby removing the value function update and lowering computational cost.
The GRPO objective in our framework is defined as follows:
\begin{align}
\mathcal{J}_{\mathrm{GRPO}}(\theta)
&= \mathbb{E}\left[ q \sim P(Q), \{o_i\}_{i=1}^G \sim \pi_{\theta_{old}}(O \mid q) \right] \nonumber \\
&\quad \frac{1}{G} \sum_{i=1}^G \left( 
\min\left( 
\rho_i A_i,\,
\mathrm{clip}\left( 
\rho_i,\,
1 - \epsilon,\,
1 + \epsilon 
\right) A_i 
\right) 
- \beta\, \mathbb{D}_{\mathrm{KL}}(\pi_\theta \| \pi_{\mathrm{ref}})
\right),
\label{eq:grpo_obj}
\end{align}
\[
\text{where} \quad
\rho_i = \frac{\pi_\theta(o_i \mid q)}{\pi_{\theta_{old}}(o_i \mid q)},
\quad
A_i = \frac{r_i - \mathrm{mean}(\{r_1, r_2, \cdots, r_G\})}
            {\mathrm{std}(\{r_1, r_2, \cdots, r_G\})}.
\]

Specifically, for each question \(q\) in the training set, 
we sample a group of outputs \(\{o_1, o_2, \ldots, o_G\}\) from the old policy \(\pi_{\theta_{old}}\) using high temperature decoding, 
and compute the group-wise relative advantage \(A_i\) for each output. 
A clipped surrogate objective, along with a KL divergence regularizer toward the reference policy \(\pi_{\mathrm{ref}}\),
is then used to update model parameters and ensure training stability. 

\subsection{Reward Design}
\label{sec:reward_design}
The reward function plays a central role in guiding and stabilizing model optimization.
In GUI action prediction, each answer is a structured action comprising a verifiable action type and associated parameters, making the task well-suited to a rule-based reward formulation.
Following DeepSeek-R1~\citep{deepseekai2025deepseekr1incentivizingreasoningcapability}, we adopt a rule-based reward \(R\) tailored for GUI tasks, which combines a format reward \(R_f\) to enforce output consistency and an action reward \(R_a\) to evaluate action correctness:
\begin{equation}
R = R_f + R_a
\end{equation}

\textbf{Format reward.} 
To ensure that model outputs are well-structured and easy to parse,
\sys requires the final answer to be enclosed in \texttt{<answer>...</answer>} tags during training.
Predictions conforming to the expected format receive \(R_f = 1\); otherwise, \(R_f = 0\).
Unlike prior methods~\citep{lu2025uir1enhancingactionprediction,liu2025infiguir1advancingmultimodalgui,luo2025guir1generalistr1style}, \sys omits explicit reasoning traces in outputs, eliminating reasoning token generation and substantially reduces training time.
For example, training Qwen2.5-VL-7B on 2K \(K\)-step GUI Transition samples requires only 9 hours, compared to 17 hours with reasoning traces,
without compromising downstream performance, as shown in Table~\ref{tab:task-type-comparison-with-2k-samples}.

\textbf{Action reward.}
We adopt a unified action space of eight types for both training and inference.
The action space comprises eight types, each as a JSON object with \textit{action\_type} and type-specific parameters:
\textit{click} and \textit{long\_press} require a target point; \textit{scroll} requires a direction; \textit{open\_app} requires an app name; \textit{input\_text} requires the input content; and \textit{navigate\_back}, \textit{navigate\_home} and \textit{wait} require no parameters.
The action reward \(R_a\) is defined accordingly:
\begin{equation}
    R_a =
    \begin{cases}
        1, & \text{if } x_1 \le \hat{x} \le x_2 \text{ and } y_1 \le \hat{y} \le y_2, \quad t \in \{\textit{click}, \textit{long\_press}\}; \\[4pt]
        1, & \text{if } \hat{t}=t \text{ and } \hat{p}=p, \quad t \in \{\textit{open\_app}, \textit{input\_text}, \textit{scroll}\}; \\[4pt]
        1, & \text{if } \hat{t}=t, \quad t \in \{\textit{navigate\_back}, \textit{navigate\_home}, \textit{wait}\}; \\[4pt]
        0, & \text{otherwise}.
    \end{cases}
\label{eq:action_reward}
\end{equation}
Here, $\hat{t}$ and $\hat{p}$ denote the predicted action type and parameter, $t$ and $p$ denote their ground-truth counterparts; 
$\hat{x}, \hat{y}$ are the predicted coordinates, and $[x_1, y_1, x_2, y_2]$ is the ground-truth bounding box.

\subsection{\sys Framework}
\textbf{\(K\)-step GUI Transition.}
While existing VLMs can parse individual GUI screens due to exposure to GUI data during pretraining, they still lack the temporal reasoning capabilities required for complex multi-step GUI tasks.
To bridge this gap, we propose the \(K\)-step GUI Transition task, which asks the model to predict the first action that transitions a given state \(S_t\) to a future state \(S_{t+k}\), as shown in Figure~\ref{fig:overview}.
Compared to annotated approaches, our task offers two key advantages.
First, while annotated tasks require costly and error-prone textual annotations for each step, \(K\)-step GUI Transition leverages state pairs directly extracted from GUI trajectories.
The future state \(S_{t+k}\), obtained after executing \(k\) actions from \(S_t\), serves as an explicit visual goal, providing a supervision signal that is not only annotation-free but also more concrete and informative than textual instructions.
Second, rather than mapping textual instructions to actions, our task compels the model to interpret and compare both the current and target GUI states, infer the transition goal, and identify the action that initiates the state change. 
Overall, by leveraging visual goals and requiring temporal reasoning across state pairs, this more challenging formulation fosters a deeper understanding of GUI dynamics and provides a scalable, practical solution for robust GUI agent training.

\textbf{Self-supervised RL.}
During training, for each sample, the model generates \(N\) candidate actions (\(N=8\) in our experiments), each evaluated by a rule-based reward that integrates format and action correctness, as detailed in Section~\ref{sec:reward_design}.
Group-wise normalized advantages are then computed, and optimization proceeds as outlined in Section~\ref{sec:preliminaries}.
GRPO is particularly well-suited to \sys for three reasons:
(1) Compared to PPO, it eliminates the need for a separate value function, typically in the same size as the policy model. This substantially reduces computational overhead and better supports our efficiency objectives;
(2) Compared to SFT, it enables flexible reward assignment tailored to each action type. For instance, click actions are considered correct if the predicted point falls within the ground-truth bounding box rather than requiring an exact match in SFT, which better reflects practical GUI grounding requirements;
(3) The \(N\)-candidate sampling mechanism encourages exploration and model can learn from optimal candidates while avoiding suboptimal ones.

\textbf{Data filtering pipeline.}
To prepare high-quality \(K\)-step GUI Transition data, we perform data filtering using the same action sampling and scoring mechanism as in training.
First, for each \(k \in \{1,2,3,4\}\), we construct a pool of candidate state pairs \((S_t, S_{t+k})\) from the original dataset.
Next, for each pair, the model generates \(8\) responses using the same sampling temperature as in GRPO training.
Each response is then evaluated with the reward function described in Section~\ref{sec:reward_design}.
Finally, we retain only those pairs with both correct and incorrect responses.
By applying this filtering process to each model independently, the final training set is both challenging and informative, and well aligned with the model's learning capacity.

Taken together, these design choices enable \sys to provide higher efficiency from three aspects:
(1) \textit{Scalable data construction.} Without relying on annotated instructions, \sys enables large-scale filtering of training data at minimal cost. For example, for Mimo-VL-7B-RL, we filtered 2,920 high-quality samples out of 8K original 1-step GUI Transition pairs, without any annotation waste.
(2) \textit{Maximized data utilization.} For each \(k\), an \(n\)-image trajectory can yield up to \(n-k\) training pairs, maximizing data utilization for fixed-length GUI trajectories.
(3) \textit{Reduced training cost.} Without explicit reasoning traces during training, \sys avoids extra token decoding and reduces training time by nearly 50\%, from 17 to 9 hours on 2K samples under our experimental setup.

%% file: sections/experiments.tex
\section{Experiments}
\label{sec:experiments}
In this section, we first detail the experimental setup, including data construction and training configurations (Section~\ref{sec:experimental_setup}). 
We then present results for models trained with \(K\)-step GUI Transition (\(k\in\{1,2,3,4\}\)), emphasizing improvements over base models and comparisons with existing baselines on GUI task automation and grounding benchmarks (Section~\ref{sec:main_results}).
To further verify the design choices in \sys, we conduct comprehensive ablation studies from four perspectives: data filtering, task formulation, reasoning configurations during RL, and training algorithms (Section~\ref{sec:ablation}).

\subsection{Experimental Setup}
\label{sec:experimental_setup}
\textbf{Training configurations.}
Using the open-source VLM-R1~\citep{shen2025vlmr1} framework, we fine-tune Qwen2.5-VL-7B~\citep{bai2025qwen25vltechnicalreport}, InternVL3-8B~\citep{chen2024internvlscalingvisionfoundation}, MimoVL-7B-SFT, and MimoVL-7B-RL~\citep{coreteam2025mimovltechnicalreport} with the pipeline described in Section~\ref{sec:framework}.
During training, only the language model is optimized while the vision encoder and projector are frozen.
All experiments are conducted on 8$\times$NVIDIA H100 GPUs.
Hyper-parameters are listed in Appendix~\ref{appendix:hyperparameters}.

\textbf{Data construction.}
All training data are sourced from the training set of AndroidControl~\citep{li2024effectsdatascaleui}, which provides GUI trajectories paired with human-labeled instructions.
These instructions enable both self-supervised \sys training and comparisons with VLMs trained using SFT (see Section~\ref{sec:ablation} for a comparison of the two training approaches).
Following the data filtering pipeline discussed in Section~\ref{sec:framework}, we select 2K samples per \(k\) for each model.
For Qwen2.5-VL-7B, the proportion of samples with either entirely correct or entirely incorrect actions was exceptionally high.
As a result, we use unfiltered data for its training.

\subsection{Benchmarks and Results}
\label{sec:main_results}
\textbf{GUI task automation.}
We evaluate \sys on two task automation benchmarks: \textit{AndroidControl}~\citep{li2024effectsdatascaleui} and \textit{GUI Odyssey}~\citep{lu2024guiodyssey}.
AndroidControl provides two test settings: \textit{AndroidControl-Low}, which assesses step-level instruction following ability (e.g., ``Type the product name in the search box''), 
and \textit{AndroidControl-High}, which evaluates long-horizon task planning (e.g., ``View the detail page of the product'').
GUI Odyssey offers a more challenging evaluation, encompassing both phone and tablet applications as well as cross-app scenarios.
The test set includes 9,134 samples in AndroidControl and 27,493 in GUI Odyssey, covering six action types: 
\textit{click}, \textit{long\_press}, \textit{scroll}, \textit{navigate\_back}, \textit{navigate\_home}, and \textit{input\_text}.
We report \textit{type match} (TM) that represents the proportion of samples with the correct action type, and \textit{exact match} (EM), which requires both the action type and all parameters to be correct.
Metrics are computed using AgentCPM-GUI~\citep{zhang2025agentcpmgui} and GUIEvalKit~\citep{xiaomi-guievalkit2025}.

\input{tables/main-task-auto.tex}

\noindent$\bullet$ \textit{\sys generally improves performance over base models on GUI task automation benchmarks.}
Table~\ref{tab:main-task-auto} presents the results of \sys, together with comparisons against both base models and models trained with annotations.
\sys achieves notable gains across all four models, especially on AndroidControl-High.
\sys-Qwen raises EM by 11.2\% over Qwen2.5-VL-7B, while \sys-Mimo-SFT and \sys-Mimo-RL reach gains of 10.3\% and 7.1\%, respectively.
On GUI Odyssey, minor declines for \sys-Mimo-SFT and \sys-Mimo-RL likely result from the 1,381 tablet episodes in the test set, whose GUI layouts differ substantially from smartphones.
Compared to models trained with annotations, \sys achieves comparable or even superior results on all benchmarks using only 2K \(K\)-step GUI Transition samples.
Overall, these results underscore the robustness and effectiveness of our approach in GUI task automation.

\textbf{GUI grounding.} 
We evaluate \sys on two GUI grounding benchmarks: \textit{ScreenSpot-v2}~\citep{wu2024osatlasfoundationactionmodel} with 1,272 samples from mobile, desktop, and web platforms, and \textit{ScreenSpot-Pro}~\citep{li2025screenspotproguigroundingprofessional} with 1,581 high-resolution screenshots for fine-grained evaluation.
Evaluations for the base models and \sys are adapted from ScreenSpot-Pro-GUI-Grounding~\citep{li2025screenspotpro}.

\noindent$\bullet$ \textit{\sys consistently outperforms base models and surpasses most existing baselines on GUI grounding benchmarks.}
Table~\ref{tab:main-grounding} summarizes the overall and baseline results.
Across all models, \sys delivers improved accuracy over base models, with the best variants reaching 2.5\% and 1.5\% gains on ScreenSpot-v2 and ScreenSpot-Pro, respectively.
Moreover, \sys surpasses all annotation-trained models except UI-Venus-Ground-7B, which is trained specifically for the GUI grounding using 107K annotated samples.
These results demonstrate that models trained solely on unlabeled GUI Transition data can effectively transfer to challenging GUI grounding tasks.
\footnote{Detailed results for different \(k\) values are provided for GUI task automation and GUI grounding benchmarks in Appendix~\ref{appendix:k-value-task-auto-details} and Appendix~\ref{appendix:k-value-grounding-details}, respectively.}
\input{tables/main-grounding.tex}

%% file: tables/main-task-auto.tex
\begin{table}[ht]
    \centering
    \caption{
      Performance comparison on GUI task automation benchmarks: AndroidControl (AC-Low, AC-High) and GUI Odyssey.
      \sys achieves substantial improvements over base models.
      \textbf{Bold}: the best result; \underline{underlined}: the second best result.
      TM: type match; EM: exact match.
      }
    \vspace{3pt}
    \setlength{\tabcolsep}{2pt}
    \label{tab:main-task-auto}
    \resizebox{\textwidth}{!}{%
    \begin{tabular}{@{}lccccccc@{}}
    \toprule
    \rowcolor[HTML]{FFFFFF} 
    \cellcolor[HTML]{FFFFFF} &
      \cellcolor[HTML]{FFFFFF} &
      \multicolumn{2}{c}{\cellcolor[HTML]{FFFFFF}\textbf{AC-Low}} &
      \multicolumn{2}{c}{\cellcolor[HTML]{FFFFFF}\textbf{AC-High}} &
      \multicolumn{2}{c}{\cellcolor[HTML]{FFFFFF}\textbf{GUI Odyssey}} \\ \cmidrule(lr){3-4} \cmidrule(lr){5-6} \cmidrule(lr){7-8} 
    \rowcolor[HTML]{FFFFFF} 
    \multirow{-2}{*}{\cellcolor[HTML]{FFFFFF}\textbf{Model}} &
      \multirow{-2}{*}{\cellcolor[HTML]{FFFFFF}\textbf{\makecell{\# Training\\Samples}}} &
      TM &
      EM &
      TM &
      EM &
      TM &
      EM \\ \midrule
    \rowcolor[HTML]{F3F3F3} 
    \multicolumn{8}{c}{\cellcolor[HTML]{F3F3F3}\textit{Proprietary models}} \\
    \rowcolor[HTML]{FFFFFF} 
    GPT-4o~\citep{hurst2024gpt} &
      - &
      74.3 &
      19.4 &
      66.3 &
      20.8 &
      34.3 &
      3.3 \\ \midrule
    \rowcolor[HTML]{F3F3F3} 
    \multicolumn{8}{c}{\cellcolor[HTML]{F3F3F3}\textit{Models trained with annotations}} \\
    \rowcolor[HTML]{FFFFFF} 
    SeeClick~\citep{cheng2024seeclickharnessingguigrounding} &
      1M &
      93.0 &
      75.0 &
      82.9 &
      59.1 &
      \cellcolor[HTML]{FFFFFF}71.0 &
      \cellcolor[HTML]{FFFFFF}53.9 \\
    \rowcolor[HTML]{FFFFFF} 
    OS-Atlas-7B~\citep{wu2024osatlasfoundationactionmodel} &
      2.3M &
      93.6 &
      85.2 &
      85.2 &
      71.2 &
      \cellcolor[HTML]{FFFFFF}- &
      \cellcolor[HTML]{FFFFFF}62.0 \\
    \rowcolor[HTML]{FFFFFF} 
    Aguvis-7B~\citep{xu2024aguvis} &
      1M &
      - &
      80.5 &
      - &
      61.5 &
      \cellcolor[HTML]{FFFFFF}- &
      \cellcolor[HTML]{FFFFFF}- \\
    \rowcolor[HTML]{FFFFFF} 
    UI-TARS-7B~\citep{qin2025uitarspioneeringautomatedgui} &
      - &
      98.0 &
      90.8 &
      83.7 &
      72.5 &
      \cellcolor[HTML]{FFFFFF}\textbf{94.6} &
      \cellcolor[HTML]{FFFFFF}\textbf{87.0} \\
    \rowcolor[HTML]{FFFFFF} 
    UI-R1-3B~\citep{lu2025uir1enhancingactionprediction} &
      136 &
      94.3 &
      88.5 &
      57.9 &
      45.4 &
      \cellcolor[HTML]{FFFFFF}52.2 &
      \cellcolor[HTML]{FFFFFF}32.5 \\
    \rowcolor[HTML]{FFFFFF} 
    GUI-R1-7B~\citep{luo2025guir1generalistr1style} &
      3K &
      85.2 &
      66.5 &
      71.6 &
      51.7 &
      \cellcolor[HTML]{FFFFFF}65.5 &
      \cellcolor[HTML]{FFFFFF}38.8 \\
    \rowcolor[HTML]{FFFFFF} 
    InfiGUI-R1-3B~\citep{liu2025infiguir1advancingmultimodalgui} &
      32K &
      96.0 &
      92.1 &
      82.7 &
      71.1 &
      \cellcolor[HTML]{FFFFFF}- &
      \cellcolor[HTML]{FFFFFF}- \\
    \rowcolor[HTML]{FFFFFF} 
    AgentCPM-GUI~\citep{liu2025infiguir1advancingmultimodalgui} &
      470K &
      94.4 &
      90.2 &
      77.7 &
      69.2 &
      \cellcolor[HTML]{FFFFFF}{\ul 90.9} &
      \cellcolor[HTML]{FFFFFF}{\ul 75.0} \\
    \rowcolor[HTML]{FFFFFF} 
    UI-Venus-Navi-7B~\citep{gu2025uivenus} &
      350K &
      97.1 &
      {\ul 92.4} &
      86.5 &
      \textbf{76.1} &
      \cellcolor[HTML]{FFFFFF}87.3 &
      \cellcolor[HTML]{FFFFFF}71.5 \\ \midrule
    \rowcolor[HTML]{F3F3F3} 
    \multicolumn{8}{c}{\cellcolor[HTML]{CBDCEB}\textit{Ours: Qwen2.5-VL-7B as the base model}} \\
    \rowcolor[HTML]{FFFFFF} 
    Qwen2.5-VL-7B~\citep{bai2025qwen25vltechnicalreport} &
      - &
      94.9 &
      83.8 &
      72.9 &
      59.2 &
      59.8 &
      44.9 \\
    \rowcolor[HTML]{FFFFFF} 
    \sys-Qwen (\(k=1\)) &
      2K &
      98.0\deltapos{3.1} &
      90.6\deltapos{6.8} &
      85.9\deltapos{13.0} &
      70.4\deltapos{11.2} &
      78.5\deltapos{18.7} &
      54.8\deltapos{9.9} \\ \midrule
    \rowcolor[HTML]{FFFFFF} 
    \rowcolor[HTML]{F3F3F3} 
    \multicolumn{8}{c}{\cellcolor[HTML]{E7F2EF}\textit{Ours: InternVL3-8B as the base model}} \\
    InternVL3-8B~\citep{chen2024internvlscalingvisionfoundation} &
      - &
      97.8 &
      90.0 &
      71.5 &
      49.8 &
      48.8 &
      20.3 \\
    \rowcolor[HTML]{FFFFFF} 
    \sys-Intern (\(k=4\)) &
      2K &
      97.3\deltaneg{0.5} & 
      88.0\deltaneg{2.0} & 
      78.5\deltapos{7.0} & 
      56.6\deltapos{6.8} &
      59.6\deltapos{10.8} & 
      23.3\deltapos{3.0} \\ \midrule

    \rowcolor[HTML]{F3F3F3} 
    \multicolumn{8}{c}{\cellcolor[HTML]{EFF5D2}\textit{Ours: Mimo-VL-7B-SFT as the base model}} \\
    \rowcolor[HTML]{FFFFFF} 
    Mimo-VL-7B-SFT~\citep{coreteam2025mimovltechnicalreport} &
      - &
      90.8 &
      85.7 &
      75.2 &
      63.1 &
      86.9 &
      62.0 \\
      \rowcolor[HTML]{FFFFFF} 
      \sys-Mimo-SFT  (\(k=3\)) &
      2K
      & \underline{98.6}\deltapos{7.8}  & \textbf{93.2}\deltapos{7.5}
      & \textbf{87.2}\deltapos{12.0} & \underline{73.4}\deltapos{10.3}
      & 86.1\deltaneg{0.8}  & 60.7\deltaneg{1.3} \\ \midrule
    \rowcolor[HTML]{F3F3F3} 
    \multicolumn{8}{c}{\cellcolor[HTML]{FFFBDE}\textit{Ours: Mimo-VL-7B-RL as the base model}} \\
    \rowcolor[HTML]{FFFFFF} 
    Mimo-VL-7B-RL~\citep{coreteam2025mimovltechnicalreport} &
      - &
      91.8 &
      87.2 &
      76.5 &
      64.6 &
      87.2 &
      63.1 \\
    \rowcolor[HTML]{FFFFFF} 
      \sys-Mimo-RL (\(k=1\)) & 2K
      & \textbf{98.9}\deltapos{7.1}  & \textbf{93.2}\deltapos{6.0}
      & \underline{86.9}\deltapos{10.4}  & 71.7\deltapos{7.1}
      & 84.8\deltaneg{2.4}  & 59.5\deltaneg{3.6} \\
      \bottomrule
    \end{tabular}
    }
    \end{table}

%% file: tables/main-grounding.tex
\begin{table}[ht]
    \centering
    \caption{
      Performance comparison on GUI grounding benchmarks: ScreenSpot-v2 and ScreenSpot-Pro.
      \sys exhibits strong generalization and achieves the second best result on ScreenSpot-Pro.
      \textbf{Bold}: the best result; \underline{underlined}: the second best result.
      }
    \vspace{3pt}
    \setlength{\tabcolsep}{4pt}
    \label{tab:main-grounding}
    \resizebox{\textwidth}{!}{%
    \begin{tabular}{@{}
    >{\columncolor[HTML]{FFFFFF}}l 
    >{\columncolor[HTML]{FFFFFF}}c 
    >{\columncolor[HTML]{FFFFFF}}c 
    >{\columncolor[HTML]{FFFFFF}}c 
    >{\columncolor[HTML]{FFFFFF}}c 
    >{\columncolor[HTML]{FFFFFF}}c 
    >{\columncolor[HTML]{FFFFFF}}c 
    >{\columncolor[HTML]{FFFFFF}}c 
    >{\columncolor[HTML]{FFFFFF}}c 
    >{\columncolor[HTML]{FFFFFF}}c @{}}
    \toprule
    \cellcolor[HTML]{FFFFFF} &
      \cellcolor[HTML]{FFFFFF} &
      \multicolumn{7}{c}{\cellcolor[HTML]{FFFFFF}\textbf{ScreenSpot-v2}} &
      \textbf{-Pro} \\ \cmidrule(lr){3-9} \cmidrule(l){10-10}
    \cellcolor[HTML]{FFFFFF} &
      \cellcolor[HTML]{FFFFFF} &
      \multicolumn{2}{c}{\cellcolor[HTML]{FFFFFF}Mobile} &
      \multicolumn{2}{c}{\cellcolor[HTML]{FFFFFF}Desktop} &
      \multicolumn{2}{c}{\cellcolor[HTML]{FFFFFF}Web} &
      \cellcolor[HTML]{FFFFFF} &
      \cellcolor[HTML]{FFFFFF} \\ \cmidrule(l){3-4} \cmidrule(l){5-6} \cmidrule(l){7-8}
    \multirow{-3}{*}{\cellcolor[HTML]{FFFFFF}\textbf{Model}} &
      \multirow{-3}{*}{\cellcolor[HTML]{FFFFFF}\textbf{\begin{tabular}[c]{@{}c@{}}\# Training \\ Samples\end{tabular}}} &
      Text &
      Icon &
      Text &
      Icon &
      Text &
      Icon &
      \multirow{-2}{*}{\cellcolor[HTML]{FFFFFF}Avg.} &
      \multirow{-2}{*}{\cellcolor[HTML]{FFFFFF}Avg.} \\ \midrule
    \multicolumn{10}{c}{\cellcolor[HTML]{F3F3F3}\textit{Models trained with annotations}}                                                                            \\
    CogAgent-18B~\citep{wang2023cogvlm}         & -    & -             & -             & -             & -             & -             & -             & -             & 7.7               \\
    SeeClick-9.6B~\citep{cheng2024seeclickharnessingguigrounding}        & 1M   & 78.4          & 50.7          & 70.1          & 29.3          & 55.2          & 32.5          & 55.1          & 1.1               \\
    UGround-7B~\citep{gou2025navigatingdigitalworldhumans}         & 1.3M & 75.1          & 84.5          & 85.1          & 61.4          & 84.6          & 71.9          & 76.3          & 16.5              \\
    OS-Atlas-7B~\citep{wu2024osatlasfoundationactionmodel}          & 2.3M & 95.2          & 75.8          & 90.7          & 63.6          & 90.6          & 77.3          & 84.1          & 18.9              \\
    ShowUI-2B~\citep{lin2024showui}            & 256K & -             & -             & -             & -             & -             & -             & -             & 7.7               \\
    UI-TARS-7B~\citep{qin2025uitarspioneeringautomatedgui}           & -    & 96.9          & 89.1          & {\ul 95.4}    & 85.0          & 93.6          & 85.2          & {\ul 91.6}    & 35.7              \\
    UI-R1-E-3B~\citep{lu2025uir1enhancingactionprediction}           & 2K   & 83.0          & \textbf{97.1} & 85.0          & \textbf{91.7} & 77.9          & \textbf{95.4} & 89.5          & 33.5              \\
    InfiGUI-R1-3B~\citep{liu2025infiguir1advancingmultimodalgui}        & 32K  & -             & -             & -             & -             & -             & -             & -             & 35.7              \\
    LPO~\citep{tang2025lpo}                  & -    & 97.9          & 82.9          & \textbf{95.9} & 86.4          & {\ul 95.6}    & 84.2          & 90.5          & -                 \\
    UI-Venus-Ground-7B~\citep{gu2025uivenus}   & 107K & \textbf{99.0}          & {\ul 90.0}    & 97.0          & {\ul 90.7}    & \textbf{96.2} & {\ul 88.7}    & \textbf{94.1} & \textbf{50.8}     \\ \midrule
    \multicolumn{10}{c}{\cellcolor[HTML]{CBDCEB}\textit{Ours: Qwen2.5-VL-7B as the base model}}                                                                     \\
    Qwen2.5-VL-7B~\citep{bai2025qwen25vltechnicalreport}        & -    & 98.3    & 86.3          & 88.7          & 67.1          & 92.7          & 81.8          & 87.7          & 26.4              \\
    \sys-Qwen (\(k=4\))     & 2K   & {\ul 98.6}          & 89.6    & 86.1          & 75.0          & 92.7          & 82.8          & 89.0\deltapos{1.3}   & 27.1\deltapos{0.7}       \\ \midrule
    \multicolumn{10}{c}{\cellcolor[HTML]{E7F2EF}\textit{Ours: InternVL3-8B as the base model}}                                                                      \\
    InternVL3-8B~\citep{chen2024internvlscalingvisionfoundation}         & -    & 93.4          & 81.5          & 80.4          & 52.1          & 91.0          & 73.4          & 81.3          & 15.0              \\
    \sys-Intern (\(k=1\))  & 2K   & 93.8 & 83.4  & 80.4 & 51.4  & 91.0   & 73.4  & 81.6\deltapos{0.3} & 15.4\deltapos{0.4}\\ \midrule

    \multicolumn{10}{c}{\cellcolor[HTML]{EFF5D2}\textit{Ours: Mimo-VL-7B-SFT as the base model}}                                                                    \\
    Mimo-VL-7B-SFT~\citep{coreteam2025mimovltechnicalreport}       & -    & 96.6          & 84.4          & 92.8          & 80.0          & 88.9          & 76.8          & 87.6          & 39.8              \\
    \sys-Mimo-SFT (\(k=1\)) & 2K & 98.3 & 87.7 & 92.3 & 82.1    & 94.0   & 79.8 & 90.1\deltapos{2.5} &40.7\deltapos{0.9}\\\midrule
    \multicolumn{10}{c}{\cellcolor[HTML]{FFFBDE}\textit{Ours: Mimo-VL-7B-RL as the base model}}                                                                     \\
    Mimo-VL-7B-RL~\citep{coreteam2025mimovltechnicalreport}        & -    & {\ul 98.3}    & 86.3          & 90.2          & 80.7          & 92.7          & 75.4          & 88.4          & 40.2              \\
    \sys-Mimo-RL (\(k=1\))  & 2K & \textbf{99.0}   & 87.7 & 91.2 & 83.6   & 89.7 & 72.9 & 88.4\deltapos{0.0}& {\ul 41.7}\deltapos{1.5} \\\bottomrule

    \end{tabular}
    }
    \end{table}

%% file: sections/ablation.tex
\subsection{Ablation Study}
\label{sec:ablation}

\input{figs/data_filter.tex}
\textbf{Data filtering.}
We evaluate InternVL3-8B, MimoVL-7B-SFT, and MimoVL-7B-RL trained with and without data filtering on both GUI task automation and GUI grounding benchmarks. 
For each \(K\)-step GUI Transition sample, each model generates \(8\) responses to build its candidate pool, retaining only those samples where predictions are both correct and incorrect.
We select 2K training samples per model and \(k\) from each filtered pool and the original training set, respectively.

\noindent$\bullet$ \textit{Data filtering improves accuracy on both GUI task automation and GUI grounding benchmarks.}
Figure~\ref{fig:data_filter} shows that models trained with filtered data achieve higher accuracy than those trained on unfiltered data in most cases.
For example, on AndroidControl-Low, Mimo-VL-7B-SFT achieves up to 4.8\% higher accuracy (Figure~\ref{fig:data_filter}(f), k=3), and on ScreenSpot-v2, up to 2.3\% (Figure~\ref{fig:data_filter}(i), k=1).
These results suggest that our filtering mechanism effectively selects more informative and challenging samples for GUI agent training.
Moreover, since \(K\)-step GUI Transition does not require human-annotated instructions, this filtering process scales easily and incurs minimal cost.

\textbf{Task formulation.}
We compare \(K\)-step GUI Transition with two annotated baselines.
To ensure fairness, we do not apply data filtering, and all models in each comparison are trained on the same set of 2K samples, with identical current states \(S_t\) and ground-truth actions. 
The only difference is the instruction type:
baselines pair \(S_t\) with a human-annotated \textit{task instruction} or \textit{step instruction},
while \(K\)-step GUI Transition uses the target state \(S_{t+k}\) as the visual instruction. 

\noindent$\bullet$ \textit{Using \(S_{t+k}\) as the visual target outperforms using textual instructions as input.}
Table~\ref{tab:task-type-comparison-with-2k-samples} shows that VLMs trained with \(K\)-step GUI Transition achieves better performance than those with annotated tasks in most cases.
For example, on AC-Low and AC-High, InternVL3-8B trained with GUI Transition achieves 4.0\% and 3.6\% higher EM accuracy, respectively, than when trained with task instructions.
Qwen2.5-VL-7B also achieves the highest EM accuracy with GUI Transition across all benchmarks.
These results indicate that \(S_{t+k}\) provides a more informative signal than human-annotated instructions.

\input{tables/task_type_comparison.tex}

\textbf{Reasoning configurations.}
To verify the effect of reasoning during training, we compare models fine-tuned on 2K \(K\)-step GUI Transition data with and without \texttt{<think>...</think>}.

\noindent$\bullet$ \textit{Excluding reasoning boosts performance and efficiency for \sys.}
Table~\ref{tab:task-type-comparison-with-2k-samples} shows that omitting explicit reasoning requirements during training not only maintains but often improves performance.
For InternVL3-8B, training without reasoning achieves up to 7.9\% higher EM on AndroidControl-High; Qwen2.5-VL-7B shows consistent gains of about 2\% across benchmarks.
Moreover, removing reasoning nearly halves training time cost, reducing it from 17 to 9 hours for Qwen2.5-VL-7B and from 15 to 7 hours for InternVL3-8B.
These results confirm that excluding reasoning both improves performance and significantly enhances training efficiency.

\input{figs/algo_comp.tex}
\textbf{Training algorithms.}
For each \(k\in\{1,2,3,4\}\), we fine-tune Qwen2.5-VL-7B and InternVL3-8B on 2K identical \(K\)-step GUI Transition data using SFT or GRPO, and evaluate them on AndroidControl.

\noindent$\bullet$ \textit{GRPO is more suitable than SFT for \(K\)-step GUI Transition.}
Figure~\ref{fig:algo_comp} shows that GRPO improves accuracy in most cases, whereas SFT consistently reduces accuracy compared to both the base models and GRPO, with drops up to 65.1\% relative to GRPO (Figure~\ref{fig:algo_comp}(c), \(k=3\)).
We attribute this to its sensitivity to format mismatch between training and evaluation.
These results confirm that GRPO is a more effective choice for transferring \(K\)-step GUI Transition knowledge to GUI task automation.

%% file: figs/data_filter.tex
\begin{figure}[!t]
    \centering
    \includegraphics[width=\linewidth]{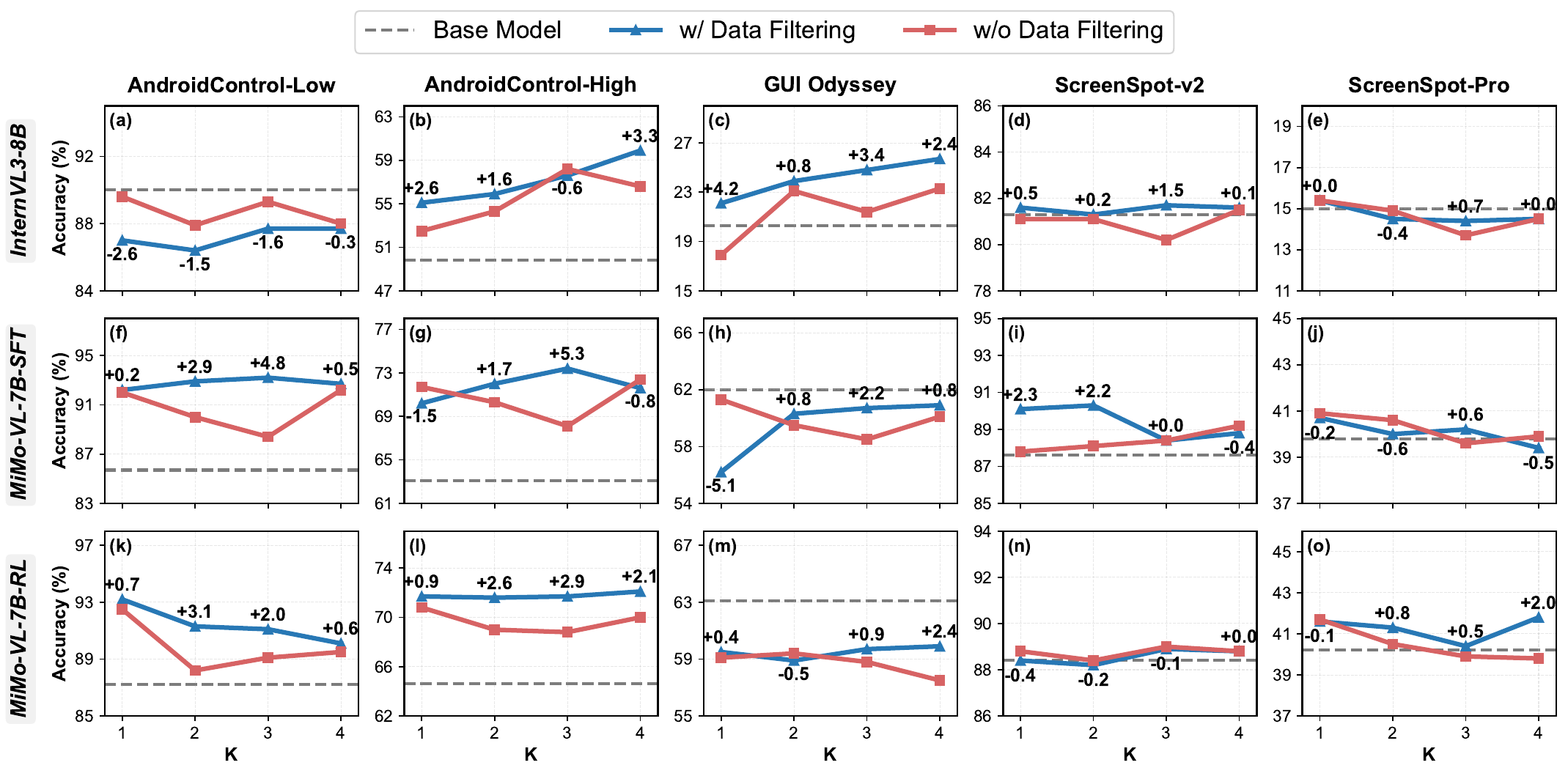}
    \caption{
      Impact of Data filtering.
      Each model is fine-tuned on 2K \(K\)-step GUI Transition samples.
      Filtered data are more informative and challenging, and outperform unfiltered ones. 
    }
    \label{fig:data_filter}
\end{figure}

%% file: tables/task_type_comparison.tex
\begin{table}[t]
  \centering
  \caption{
    Performance on GUI task automation under different training settings.
    \sys outperforms models trained with textual instructions or with explicit reasoning requirements.
    \textbf{Bold}: the best result.
    TM: type match; GR: grounding accuracy for clicks; EM: exact match.
    }
  \vspace{3pt}
  \setlength{\tabcolsep}{3pt}
  \label{tab:task-type-comparison-with-2k-samples}
  \resizebox{\textwidth}{!}{%
  \begin{tabular}{@{}lccccccccc@{}}
  \toprule
  & \multicolumn{3}{c}{\cellcolor[HTML]{FFFFFF}\textbf{AndroidControl-Low}}
    & \multicolumn{3}{c}{\cellcolor[HTML]{FFFFFF}\textbf{AndroidControl-High}}
    & \multicolumn{3}{c}{\cellcolor[HTML]{FFFFFF}\textbf{GUI Odyssey}} \\ 
  \cmidrule(l){2-4} \cmidrule(l){5-7} \cmidrule(l){8-10}
  \multirow{-2}{*}{\cellcolor[HTML]{FFFFFF}\textbf{Model}}
    & TM & GR & EM & TM & GR & EM & TM & GR & EM \\ 
    \midrule
    \multicolumn{10}{c}{\cellcolor[HTML]{CBDCEB}\textit{Base model: Qwen2.5-VL-7B}} \\
    Qwen2.5-VL-7B
      & 94.9 & 90.9 & 83.8
      & 72.9 & 66.6 & 59.2
      & 59.8 & 47.5 & 44.9 \\
    \textit{+ Task Instruction}
      & 97.9\deltapos{3.0} & 93.5\deltapos{2.6} & 90.5\deltapos{6.7}
      & 85.3\deltapos{12.4} & 76.2\deltapos{9.6} & 69.9\deltapos{10.7}
      & 74.1\deltapos{14.3} & 62.0\deltapos{14.5} & 51.8\deltapos{6.9} \\
      \textit{+ Step Instruction}
      & 97.7\deltapos{2.8} & 93.7\deltapos{2.8} & 86.4\deltapos{2.6}
      & 82.4\deltapos{9.5}  & 73.1\deltapos{6.5} & 67.2\deltapos{8.0}
      & 74.5\deltapos{14.7}& 62.7\deltapos{15.2}& 51.5\deltapos{6.6} \\
    \rowcolor[HTML]{F3F3F3}
    \textit{Ours (w/ reasoning)}
      & 95.5\deltapos{0.6} & 91.1\deltapos{0.2} & 88.2\deltapos{4.4}
      & 83.8\deltapos{10.9} & 75.6\deltapos{9.0} & 69.0\deltapos{9.8}
      & 74.0\deltapos{14.2} & 63.5\deltapos{16.0} & 51.6\deltapos{6.7} \\
    \rowcolor[HTML]{F3F3F3}
    \textbf{\textit{Ours}}
      & \textbf{98.0}\deltapos{3.1} & \textbf{94.0}\deltapos{3.1} & \textbf{90.6}\deltapos{6.8}
      & \textbf{85.9}\deltapos{13.0}& \textbf{77.5}\deltapos{10.9}& \textbf{70.4}\deltapos{11.2}
      & \textbf{78.5}\deltapos{18.7}& \textbf{67.2}\deltapos{19.7}& \textbf{54.8}\deltapos{9.9} \\ 
  \midrule
  \multicolumn{10}{c}{\cellcolor[HTML]{E7F2EF}\textit{Base model: InternVL3-8B}} \\
  InternVL3-8B
    & 97.8 & 92.4 & 90.0
    & 71.5 & 54.6 & 49.8
    & 48.8 & 20.2 & 20.3 \\
  \textit{+ Task Instruction}
    & 95.9\deltaneg{1.9} & \textbf{92.8}\deltapos{0.4} & 85.3\deltaneg{4.7}
    & 79.7\deltapos{8.2} & 65.8\deltapos{11.2} & 54.6\deltapos{4.8}
    & 57.3\deltapos{8.5} & 31.5\deltapos{11.3} & 26.5\deltapos{6.2} \\
  \textit{+ Step Instruction}
    & 96.8\deltaneg{1.0} & 92.8\deltapos{0.4} & 86.0\deltaneg{4.0}
    & \textbf{80.7}\deltapos{9.2} & \textbf{66.0}\deltapos{11.4} & 54.5\deltapos{4.7}
    & \textbf{62.6}\deltapos{13.8} & \textbf{34.4}\deltapos{14.2} & \textbf{28.8}\deltapos{8.5} \\
  \rowcolor[HTML]{F3F3F3}
    \textit{Ours (w/ reasoning)}
    & 97.3\deltaneg{0.5} & 92.3\deltapos{0.1} & 87.8\deltaneg{2.2}
    & 72.4\deltapos{0.9} & 55.8\deltapos{1.2} & 50.3\deltapos{0.5}
    & 38.9\deltaneg{9.9} & 18.8\deltaneg{1.4} & 16.2\deltaneg{4.1} \\
  \rowcolor[HTML]{F3F3F3}
    \textbf{\textit{Ours}}
      & \textbf{97.5}\deltaneg{0.3} & 92.6\deltapos{0.2} & \textbf{89.3}\deltaneg{0.7}
      & 78.6\deltapos{7.1} & 61.7\deltapos{7.1} & \textbf{58.2}\deltapos{8.4}
      & 51.4\deltapos{2.6} & 21.4\deltapos{1.2} & 21.4\deltapos{1.1} \\
  \bottomrule
  \end{tabular}%
  }
\end{table}

%% file: figs/algo_comp.tex
\begin{figure}[ht]
    \centering
    \includegraphics[width=\linewidth]{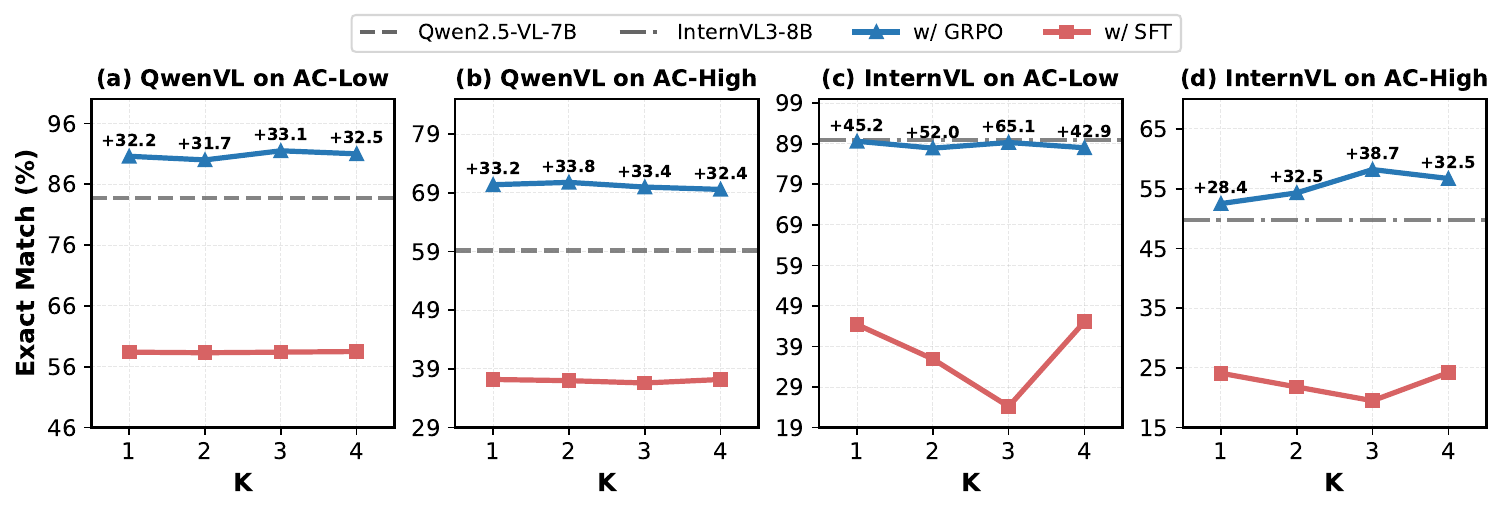}
    \vspace{-20pt}
    \caption{
        Comparison of training algorithms for the \(K\)-step GUI Transition task (\(k \in \{1,2,3,4\}\)).
        Qwen2.5-VL-7B and InternVL3-8B are fine-tuned with 2K samples for each \(k\) and evaluated on AndroidControl.
        GRPO provides notable performance gains over SFT for all models and settings.
      }
      
    \label{fig:algo_comp}
    \vspace{-3pt}
  \end{figure}
  

%% file: sections/conclusion.tex
\section{Conclusion}
\label{sec:conclusion}
This study introduces \sys, a self-supervised reinforcement learning framework for training VLM-based GUI agents without relying on costly annotations.
Based on the \(K\)-step GUI Transition training task, enhancing VLMs with \sys captures temporal dynamics between GUI states and provides a scalable, annotation-free training signal.
Experiments across four VLMs and four benchmarks show consistent improvements, including up to 11.2\% gains in GUI task automation accuracy and strong generalization to GUI grounding tasks.
These results demonstrate that self-supervised RL can effectively exploit unlabeled GUI trajectories, offering a practical and efficient alternative to training tasks with human-annotated instructions.

%% file: sections/appendix.tex
\section{Training Hyper-parameters}
\label{appendix:hyperparameters}
The hyper-parameter details for \sys are provided in Table~\ref{tab:hyperparams}.

\begin{table}[ht]
    \centering
    \caption{Hyper-parameter settings used for all GRPO training.}
    \label{tab:hyperparams}
    \begin{tabular}{ll}
    \toprule
    Hyper-parameter                    & Value                     \\
    \midrule
    learning\_rate                     & from 1e-6 to 0         \\
    temperature                        & 0.9                       \\
    num\_generations                   & 8                         \\
    num\_train\_epochs                 & 4                         \\
    max\_prompt\_length                & 1024                      \\
    max\_completion\_length            & 256                       \\
    per\_device\_train\_batch\_size    & 2                         \\
    gradient\_accumulation\_steps      & 8                         \\
    \(\epsilon\) (clipping parameter)  & 0.2                      \\
    \(\beta\) (KL coefficient)         & 0.04                      \\
    \bottomrule
    \end{tabular}
\end{table}

\newpage
\section{Task Automation Results with and without Data Filtering}
\label{appendix:k-value-task-auto-details}

Table~\ref{tab:k-value-task-auto-no-filter} and Table~\ref{tab:k-value-task-auto-with-filter} report the performance of \sys on AndroidControl and GUI Odyssey benchmarks using 2K \(K\)-step GUI Transition samples, with and without data filtering.
Without data filtering, as shown in Table~\ref{tab:k-value-task-auto-no-filter}, \sys trained with each \(k\) value outperforms the base models on most benchmarks. 
Specifically, for Qwen2.5-VL-7B, \sys delivers consistent improvements across all benchmarks;
for the other three models, \sys also improves task automation in most cases, especially on AndroidControl.
These results demonstrate that conditioning on the future state \(S_{t+k}\) consistently provides an effective supervision signal across different transition step sizes.
We apply data filtering is to InternVL3-8B, MimoVL-7B-SFT, and MimoVL-7B-RL.
With data filtering, as shown in Table~\ref{tab:k-value-task-auto-with-filter}, model performance further improves in most settings, confirming that filtering enhances data quality and strengthens VLM optimization.

\input{tables/appendix-task-auto-details-no-filter.tex}

\clearpage

\input{tables/appendix-task-auto-details-with-filter.tex}

\clearpage
\section{Grounding Results with and without Data Filtering}
\label{appendix:k-value-grounding-details}
We report the accuracy of \sys on two GUI grounding benchmarks: ScreenSpot-v2 and ScreenSpot-Pro, under both filtered and unfiltered settings.

For ScreenSpot-v2 (Table~\ref{tab:appendix-ss-v2-details}), all models except \sys-Intern achieve consistent improvements over their respective baselines without data filtering. 
With data filtering (Table~\ref{tab:appendix-ss-v2-details-with-filter}), \sys-Intern models also surpass their baselines, indicating the benefit of filtering.

For ScreenSpot-Pro (Table~\ref{tab:appendix-ss-pro-details}), all \sys-Qwen models improve without data filtering, while other models exhibit mixed results across different \(k\). 
We attribute the performance drop to the high-resolution desktop screenshots in ScreenSpot-Pro, which are not present in our training data. 
With data filtering (Table~\ref{tab:appendix-ss-pro-details-with-filter}), \sys-Mimo-RL achieves consistent gains for all \(k\).

Overall, \sys demonstrates strong generalization on GUI grounding, with performance gains in most cases, especially when data filtering is applied.

\input{tables/appendix-ss-v2-details.tex}
\input{tables/appendix-ss-v2-details-with-filter.tex}

\input{tables/appendix-ss-pro-details.tex}
\clearpage
\input{tables/appendix-ss-pro-details-with-filter.tex}

%% file: tables/appendix-task-auto-details-no-filter.tex
\begin{table}[ht]
  \centering
  \caption{
    Performance on task automation benchmarks: AndroidControl and GUI Odyssey.
    Models are fine-tuned with 2K \(k\)-step UI Transition samples for each \(k \in \{1,2,3,4\}\), \textbf{without model-specific data filtering}.
    The target states \(S_{t+k}\) with different \(k\) values provide effective visual instructions for GUI agent training.
    TM: type match; GR: grounding accuracy for clicks; EM: exact match.
  }
  \vspace{1em}
  \setlength{\tabcolsep}{2pt}
  \label{tab:k-value-task-auto-no-filter}
  \resizebox{\textwidth}{!}{%
  \begin{tabular}{@{}lccccccccc@{}}
  \toprule
  & \multicolumn{3}{c}{\textbf{AndroidControl-Low}}
    & \multicolumn{3}{c}{\textbf{AndroidControl-High}}
    & \multicolumn{3}{c}{\textbf{GUI Odyssey}} \\
  \cmidrule(l){2-4} \cmidrule(l){5-7} \cmidrule(l){8-10}
  \multirow{-2}{*}{\textbf{Model}}
    & TM & GR & EM & TM & GR & EM & TM & GR & EM \\
  \midrule
  \rowcolor[HTML]{CBDCEB}
  Qwen2.5-VL-7B
    & 94.9 & 90.9 & 83.8
    & 72.9 & 66.6 & 59.2
    & 59.8 & 47.5 & 44.9 \\
  \sys-Qwen (\(k=1\))
    & 98.0\deltapos{3.1} & 94.0\deltapos{3.1} & 90.6\deltapos{6.8}
    & 85.9\deltapos{13.0} & 77.5\deltapos{10.9} & 70.4\deltapos{11.2}
    & 78.5\deltapos{18.7} & 67.2\deltapos{19.7} & 54.8\deltapos{9.9} \\
  \sys-Qwen (\(k=2\))
    & 98.0\deltapos{3.1} & 93.7\deltapos{2.8} & 90.0\deltapos{6.2}
    & 85.9\deltapos{13.0} & 76.9\deltapos{10.3} & 70.8\deltapos{11.6}
    & 79.4\deltapos{19.6} & 68.6\deltapos{21.1} & 55.7\deltapos{10.8} \\
  \sys-Qwen (\(k=3\))
    & 97.9\deltapos{3.0} & 93.6\deltapos{2.7} & 91.5\deltapos{7.7}
    & 85.3\deltapos{12.4} & 77.7\deltapos{11.1} & 70.0\deltapos{10.8}
    & 77.1\deltapos{17.3} & 67.6\deltapos{20.1} & 53.8\deltapos{8.9} \\
  \sys-Qwen (\(k=4\))
    & 97.8\deltapos{2.9} & 92.8\deltapos{1.9} & 91.0\deltapos{7.2}
    & 84.8\deltapos{11.9} & 77.1\deltapos{10.5} & 69.6\deltapos{10.4}
    & 76.0\deltapos{16.2} & 65.7\deltapos{18.2} & 53.1\deltapos{8.2} \\
  \midrule
  \rowcolor[HTML]{E7F2EF}
  InternVL3-8B
    & 97.8 & 92.4 & 90.0
    & 71.5 & 54.6 & 49.8
    & 48.8 & 20.2 & 20.3 \\

  \sys-Intern (\(k=1\))
    & 98.1\deltapos{0.3} & 92.8\deltapos{0.4} & 89.6\deltaneg{0.4}
    & 73.1\deltapos{1.6} & 56.0\deltapos{1.4} & 52.5\deltapos{2.7}
    & 42.9\deltaneg{5.9} & 17.6\deltaneg{2.6} & 17.9\deltaneg{2.4} \\

  \sys-Intern (\(k=2\))
    & 97.6\deltaneg{0.2} & 92.9\deltapos{0.5} & 87.9\deltaneg{2.1}
    & 76.3\deltapos{4.8} & 59.8\deltapos{5.2} & 54.3\deltapos{4.5}
    & 59.4\deltapos{10.6} & 24.4\deltapos{4.2} & 23.1\deltapos{2.8} \\

  \sys-Intern (\(k=3\))
    & 97.5\deltaneg{0.3} & 92.6\deltapos{0.2} & 89.3\deltaneg{0.7}
    & 78.6\deltapos{7.1} & 61.7\deltapos{7.1} & 58.2\deltapos{8.4}
    & 51.4\deltapos{2.6} & 21.4\deltapos{1.2} & 21.4\deltapos{1.1} \\

  \sys-Intern (\(k=4\))
    & 97.3\deltaneg{0.5} & 92.9\deltapos{0.5} & 88.0\deltaneg{2.0}
    & 78.5\deltapos{7.0} & 64.3\deltapos{9.7} & 56.6\deltapos{6.8}
    & 59.6\deltapos{10.8} & 25.9\deltapos{5.7} & 23.3\deltapos{3.0} \\
  \midrule

  \rowcolor[HTML]{EFF5D2}
  Mimo-VL-7B-SFT
    & 90.8 & 93.5 & 85.7
    & 75.2 & 75.7 & 63.1
    & 86.9 & 66.3 & 62.0 \\
  \sys-Mimo-SFT (\(k=1\))
    & 98.6\deltapos{7.8} & 93.8\deltapos{0.3} & 92.0\deltapos{6.3}
    & 86.8\deltapos{11.6} & 74.3\deltaneg{1.4} & 71.7\deltapos{8.6}
    & 85.4\deltaneg{1.5} & 67.1\deltapos{0.8} & 61.3\deltaneg{0.7} \\
  \sys-Mimo-SFT (\(k=2\))
    & 98.5\deltapos{7.7} & 92.7\deltaneg{0.8} & 90.0\deltapos{4.3}
    & 87.0\deltapos{11.8} & 73.9\deltaneg{1.8} & 70.3\deltapos{7.2}
    & 85.0\deltaneg{1.9} & 66.0\deltaneg{0.3} & 59.5\deltaneg{2.5} \\
  \sys-Mimo-SFT (\(k=3\))
    & 98.2\deltapos{7.4} & 92.5\deltaneg{1.0} & 88.4\deltapos{2.7}
    & 85.5\deltapos{10.3} & 72.2\deltaneg{3.5} & 68.1\deltapos{5.0}
    & 86.0\deltaneg{0.9} & 65.4\deltaneg{0.9} & 58.5\deltaneg{3.5} \\
  \sys-Mimo-SFT (\(k=4\))
    & 98.3\deltapos{7.5} & 93.0\deltaneg{0.5} & 92.2\deltapos{6.5}
    & 86.9\deltapos{11.7} & 73.4\deltaneg{2.3} & 72.4\deltapos{9.3}
    & 85.9\deltaneg{1.0} & 67.7\deltapos{1.4} & 60.1\deltaneg{1.9} \\
    \midrule

  \rowcolor[HTML]{FFFBDE}
  Mimo-VL-7B-RL
    & 91.8 & 94.5 & 87.2
    & 76.5 & 77.5 & 64.6
    & 87.2 & 67.9 & 63.1 \\
  \sys-Mimo-RL (\(k=1\))
    & 98.7\deltapos{6.9} & 94.7\deltapos{0.2} & 92.5\deltapos{5.3}
    & 86.7\deltapos{10.2} & 77.0\deltaneg{0.5} & 70.8\deltapos{6.2}
    & 85.0\deltaneg{2.2} & 67.8\deltaneg{0.1} & 59.1\deltaneg{4.0} \\
  \sys-Mimo-RL (\(k=2\))
    & 98.7\deltapos{6.9} & 93.3\deltaneg{1.2} & 88.2\deltapos{1.0}
    & 86.1\deltapos{9.6} & 72.7\deltaneg{4.8} & 69.0\deltapos{4.4}
    & 85.3\deltaneg{1.9} & 66.2\deltaneg{1.7} & 59.4\deltaneg{3.7} \\
  \sys-Mimo-RL (\(k=3\))
    & 98.0\deltapos{6.2} & 93.8\deltaneg{0.7} & 89.1\deltapos{1.9}
    & 85.8\deltapos{9.3} & 73.6\deltaneg{3.9} & 68.8\deltapos{4.2}
    & 85.9\deltaneg{1.3} & 66.0\deltaneg{1.9} & 58.8\deltaneg{4.3} \\
  \sys-Mimo-RL (\(k=4\))
    & 98.5\deltapos{6.7} & 93.0\deltaneg{1.5} & 89.5\deltapos{2.3}
    & 86.6\deltapos{10.1} & 73.0\deltaneg{4.5} & 70.1\deltapos{5.5}
    & 85.3\deltaneg{1.9} & 66.1\deltaneg{1.8} & 57.5\deltaneg{5.6} \\
  \bottomrule
  \end{tabular}%
  }
\end{table}

%% file: tables/appendix-task-auto-details-with-filter.tex
\begin{table}[ht]
    \centering
    \caption{
      Performance on task automation benchmarks: AndroidControl and GUI Odyssey.
      Compared to training without data filtering, applying filtering yields greater improvements in most cases.
      TM: type match; GR: grounding accuracy for clicks; EM: exact match.
    }
    \vspace{1em}
    \setlength{\tabcolsep}{2pt}
    \label{tab:k-value-task-auto-with-filter}
    \resizebox{\textwidth}{!}{%
    \begin{tabular}{@{}lccccccccc@{}}
    \toprule
    & \multicolumn{3}{c}{\textbf{AndroidControl-Low}}
      & \multicolumn{3}{c}{\textbf{AndroidControl-High}}
      & \multicolumn{3}{c}{\textbf{GUI Odyssey}} \\
    \cmidrule(l){2-4} \cmidrule(l){5-7} \cmidrule(l){8-10}
    \multirow{-2}{*}{\textbf{Model}}
      & TM & GR & EM & TM & GR & EM & TM & GR & EM \\
    \midrule
    \rowcolor[HTML]{E7F2EF}
    InternVL3-8B
      & 97.8 & 92.4 & 90.0
      & 71.5 & 54.6 & 49.8
      & 48.8 & 20.2 & 20.3 \\
    \sys-Intern (\(k=1\))
      & 96.2\deltaneg{1.6} & 92.6\deltapos{0.2} & 87.0\deltaneg{3.0}
      & 78.6\deltapos{7.1} & 63.2\deltapos{8.6} & 55.1\deltapos{5.3}
      & 51.6\deltapos{2.8} & 24.5\deltapos{4.3} & 22.1\deltapos{1.8} \\
    \sys-Intern (\(k=2\))
      & 97.2\deltaneg{0.6} & 92.3\deltaneg{0.1} & 86.4\deltaneg{3.6}
      & 79.4\deltapos{7.9} & 63.3\deltapos{8.7} & 55.9\deltapos{6.1}
      & 51.9\deltapos{3.1} & 24.4\deltapos{4.2} & 23.9\deltapos{3.6} \\
    
    \sys-Intern (\(k=3\))
      & 96.3\deltaneg{1.5} & 92.7\deltapos{0.3} & 87.7\deltaneg{2.3}
      & 79.9\deltapos{8.4} & 65.5\deltapos{10.9} & 57.6\deltapos{7.8}
      & 59.5\deltapos{10.7} & 27.2\deltapos{7.0} & 24.8\deltapos{4.5} \\
    
    \sys-Intern (\(k=4\))
      & 96.2\deltaneg{1.6} & 92.8\deltapos{0.4} & 87.7\deltaneg{2.3}
      & 81.0\deltapos{9.5} & 67.8\deltapos{13.2} & 59.9\deltapos{10.1}
      & 57.0\deltapos{8.2} & 27.4\deltapos{7.2} & 25.7\deltapos{5.4} \\
    \midrule
    \rowcolor[HTML]{EFF5D2}
    Mimo-VL-7B-SFT
      & 90.8 & 93.5 & 85.7
      & 75.2 & 75.7 & 63.1
      & 86.9 & 66.3 & 62.0 \\
    \sys-Mimo-SFT (\(k=1\))
      & 97.8\deltapos{7.0} & 93.8\deltapos{0.3} & 92.2\deltapos{6.5}
      & 84.6\deltapos{9.4} & 75.7\deltapos{0.0} & 70.2\deltapos{7.1}
      & 81.8\deltaneg{5.1} & 64.3\deltaneg{2.0} & 56.2\deltaneg{5.8} \\
    
    \sys-Mimo-SFT (\(k=2\))
      & 98.6\deltapos{7.8} & 93.3\deltaneg{0.2} & 92.9\deltapos{7.2}
      & 86.4\deltapos{11.2} & 75.8\deltapos{0.1} & 72.0\deltapos{8.9}
      & 85.2\deltaneg{1.7} & 66.6\deltapos{0.3} & 60.3\deltaneg{1.7} \\
    
    \sys-Mimo-SFT (\(k=3\))
      & 98.6\deltapos{7.8} & 94.0\deltapos{0.5} & 93.2\deltapos{7.5}
      & 87.2\deltapos{12.0} & 75.6\deltaneg{0.1} & 73.4\deltapos{10.3}
      & 86.1\deltaneg{0.8} & 66.3\deltapos{0.0} & 60.7\deltaneg{1.3} \\
    
    \sys-Mimo-SFT (\(k=4\))
      & 98.6\deltapos{7.8} & 93.5\deltapos{0.0} & 92.7\deltapos{7.0}
      & 85.8\deltapos{10.6} & 73.9\deltaneg{1.8} & 71.6\deltapos{8.5}
      & 85.9\deltaneg{1.0} & 67.5\deltapos{1.2} & 60.9\deltaneg{1.1} \\
    \rowcolor[HTML]{FFFBDE}
    \midrule
    Mimo-VL-7B-RL
      & 91.8 & 94.5 & 87.2
      & 76.5 & 77.5 & 64.6
      & 87.2 & 67.9 & 63.1 \\
    \sys-Mimo-RL (\(k=1\))
      & 98.9\deltapos{7.1} & 94.3\deltaneg{0.2} & 93.2\deltapos{6.0}
      & 86.9\deltapos{10.4} & 75.9\deltaneg{1.6} & 71.7\deltapos{7.1}
      & 84.8\deltaneg{2.4} & 67.5\deltaneg{0.4} & 59.5\deltaneg{3.6} \\
    
    \sys-Mimo-RL (\(k=2\))
      & 97.7\deltapos{5.9} & 93.7\deltaneg{0.8} & 91.3\deltapos{4.1}
      & 87.6\deltapos{11.1} & 75.9\deltaneg{1.6} & 71.6\deltapos{7.0}
      & 84.9\deltaneg{2.3} & 65.4\deltaneg{2.5} & 58.9\deltaneg{4.2} \\
    
    \sys-Mimo-RL (\(k=3\))
      & 97.3\deltapos{5.5} & 93.9\deltaneg{0.6} & 91.1\deltapos{3.9}
      & 87.1\deltapos{10.6} & 77.3\deltaneg{0.2} & 71.7\deltapos{7.1}
      & 84.9\deltaneg{2.3} & 67.6\deltaneg{0.3} & 59.7\deltaneg{3.4} \\
    
    \sys-Mimo-RL (\(k=4\))
      & 96.8\deltapos{5.0} & 94.2\deltaneg{0.3} & 90.9\deltapos{3.7}
      & 87.0\deltapos{10.5} & 77.0\deltaneg{0.5} & 72.1\deltapos{7.5}
      & 84.7\deltaneg{2.5} & 68.3\deltapos{0.4} & 59.9\deltaneg{3.2} \\
    \bottomrule
    \end{tabular}%
    }
  \end{table}

%% file: tables/appendix-ss-v2-details.tex
\begin{table}[ht]
    \centering
    \caption{
      Performance on GUI grounding benchmark: ScreenSpot-v2.
      Models are fine-tuned with 2K \(k\)-step UI Transition samples for each \(k \in \{1,2,3,4\}\), \textbf{without model-specific data filtering}.
    }
    \vspace{1em}
    \label{tab:appendix-ss-v2-details}
    \resizebox{\textwidth}{!}{%
    \begin{tabular}{@{}lcccccccccc@{}}
    \toprule
     &
      \multicolumn{3}{c}{\textbf{Mobile}} &
      \multicolumn{3}{c}{\textbf{Desktop}} &
      \multicolumn{3}{c}{\textbf{Web}} &
       \\ \cmidrule(l){2-4} \cmidrule(l){5-7} \cmidrule(l){8-10}
    \multirow{-2}{*}{\textbf{Model}} &
      Text &
      Icon &
      Avg. &
      Text &
      Icon &
      Avg. &
      Text &
      Icon &
      Avg. &
      \multirow{-2}{*}{\textbf{Avg.}} \\ \midrule
    \rowcolor[HTML]{CBDCEB} 
    Qwen2.5-VL-7B       & 98.3 & 86.3 & 93.2 & 88.7 & 67.1 & 79.6 & 92.7 & 81.8 & 87.6 & 87.7        \\
    \sys-Qwen (\(k=1\))     & 98.6 & 87.7 & 94.0 & 88.1 & 71.4 & 81.1 & 92.7 & 81.8 & 87.6 & 88.4\deltapos{0.7} \\
    \sys-Qwen (\(k=2\))     & 98.6 & 89.1 & 94.6 & 87.6 & 73.6 & 81.7 & 92.3 & 80.8 & 87.0 & 88.6\deltapos{0.9} \\
    \sys-Qwen (\(k=3\))     & 99.0 & 88.6 & 94.6 & 86.1 & 72.9 & 80.5 & 92.3 & 80.8 & 87.0 & 88.3\deltapos{0.6} \\
    \sys-Qwen (\(k=4\))     & 98.6 & 89.6 & 94.8 & 86.1 & 75.0 & 81.4 & 92.7 & 82.8 & 88.1 & 89.0\deltapos{1.3} \\ \midrule
    \rowcolor[HTML]{E7F2EF}    
    InternVL3-8B         & 93.4 & 81.5 & 88.4 & 80.4 & 52.1 & 68.6 & 91   & 73.4 & 82.8 & 81.3        \\
    \sys-Intern (\(k=1\))   & 94.1 & 81.5 & 88.8 & 77.3 & 52.1 & 66.8 & 91.5 & 73.4 & 83.1 & 81.1\deltaneg{0.2} \\
    \sys-Intern (\(k=2\))   & 93.8 & 80.6 & 88.2 & 80.4 & 52.1 & 68.6 & 91.5 & 72.4 & 82.6 & 81.1\deltaneg{0.2} \\
    \sys-Intern (\(k=3\))   & 94.8 & 80.1 & 88.6 & 78.9 & 47.1 & 65.6 & 90.6 & 71.4 & 81.7 & 80.2\deltaneg{0.9} \\
    \sys-Intern (\(k=4\))   & 94.1 & 82.0 & 89.0 & 79.4 & 54.3 & 68.9 & 91.9 & 71.9 & 82.6 & 81.5\deltapos{0.2} \\ \midrule
    \rowcolor[HTML]{EFF5D2} 
    Mimo-VL-7B-SFT       & 96.6 & 84.4 & 91.4 & 92.8 & 80.0 & 87.4 & 88.9 & 76.8 & 83.3 & 87.6        \\
    \sys-Mimo-SFT (\(k=1\)) & 97.2 & 86.3 & 92.6 & 91.8 & 79.3 & 86.5 & 90.6 & 74.9 & 83.3 & 87.8\deltapos{0.2} \\
    \sys-Mimo-SFT (\(k=2\)) & 95.9 & 84.4 & 91.0 & 92.8 & 82.9 & 88.6 & 91.9 & 75.9 & 84.4 & 88.1\deltapos{0.5} \\
    \sys-Mimo-SFT (\(k=3\)) & 96.6 & 86.3 & 92.2 & 91.8 & 82.9 & 88.0 & 90.6 & 77.3 & 84.4 & 88.4\deltapos{0.8} \\
    \sys-Mimo-SFT (\(k=4\)) & 96.9 & 87.2 & 92.8 & 91.8 & 84.3 & 88.6 & 89.7 & 80.3 & 85.4 & 89.2\deltapos{1.6} \\ \midrule
    \rowcolor[HTML]{FFFBDE} 
    Mimo-VL-7B-RL        & 98.3 & 86.3 & 93.2 & 90.2 & 80.7 & 86.2 & 92.7 & 75.4 & 84.7 & 88.4        \\
    \sys-Mimo-RL (\(k=1\))  & 99.3 & 88.2 & 94.6 & 91.8 & 80.7 & 87.1 & 90.2 & 75.4 & 83.3 & 88.8\deltapos{0.4} \\
    \sys-Mimo-RL (\(k=2\))  & 97.9 & 87.2 & 93.4 & 91.8 & 81.4 & 87.4 & 91   & 74.4 & 83.3 & 88.4\deltapos{0.0} \\
    \sys-Mimo-RL (\(k=3\)) & 98.3 & 87.2 & 93.6 & 91.8 & 83.6 & 88.3 & 91   & 76.4 & 84.2 & 89.0\deltapos{0.6} \\
    \sys-Mimo-RL (\(k=4\))  & 99.3 & 87.7 & 94.4 & 91.8 & 81.4 & 87.4 & 90.6 & 75.4 & 83.5 & 88.8\deltapos{0.4} \\ \bottomrule
    \end{tabular}%
    }
    \end{table}

%% file: tables/appendix-ss-v2-details-with-filter.tex
\begin{table}[ht]
    \centering
    \caption{
      Performance on GUI grounding benchmark: ScreenSpot-v2.
      \textbf{Model-specific data filtering is applied} to InternVL3-8B, MimoVL-7B-SFT, and MimoVL-7B-RL.
      For each model, we select 2K \(k\)-step UI Transition samples for each \(k \in \{1,2,3,4\}\) from a pool of candidates.
    }
    \vspace{1em}
    \label{tab:appendix-ss-v2-details-with-filter}
    \resizebox{\textwidth}{!}{%
    \begin{tabular}{@{}lcccccccccc@{}}
    \toprule 
     &
      \multicolumn{3}{c}{\textbf{Mobile}} &
      \multicolumn{3}{c}{\textbf{Desktop}} &
      \multicolumn{3}{c}{\textbf{Web}} &
       \\ \cmidrule(l){2-4} \cmidrule(l){5-7} \cmidrule(l){8-10}
     
    \multirow{-2}{*}{\textbf{Model}} &
      Text &
      Icon &
      Avg. &
      Text &
      Icon &
      Avg. &
      Text &
      Icon &
      Avg. &
      \multirow{-2}{*}{\textbf{Avg.}} \\ \midrule
    \rowcolor[HTML]{E7F2EF} 
    InternVL3-8B         & 93.4 & 81.5 & 88.4 & 80.4 & 52.1 & 68.6 & 91.0   & 73.4 & 82.8 & 81.3        \\
    \sys-Intern (\(k=1\))   & 93.8 & 83.4 & 89.4 & 80.4 & 51.4 & 68.3 & 91.0   & 73.4 & 82.8 & 81.6\deltapos{0.3} \\
    \sys-Intern (\(k=2\))   & 94.5 & 82   & 89.2 & 78.9 & 51.4 & 67.4 & 91.9 & 72.4 & 82.8 & 81.3\deltapos{0.0} \\
    \sys-Intern (\(k=3\))   & 94.5 & 83.4 & 89.8 & 79.4 & 50.7 & 67.4 & 91.9 & 73.4 & 83.3 & 81.7\deltapos{0.4} \\
    \sys-Intern (\(k=4\))  & 93.4 & 83.4 & 89.2 & 77.8 & 54.3 & 68.0   & 91.5 & 73.9 & 83.3 & 81.6\deltapos{0.3} \\ \midrule
    \rowcolor[HTML]{EFF5D2} 
    Mimo-VL-7B-SFT       & 96.6 & 84.4 & 91.4 & 92.8 & 80.0 & 87.4 & 88.9 & 76.8 & 83.3 & 87.6        \\
    \sys-Mimo-SFT (\(k=1\)) & 98.3 & 87.7 & 93.8 & 92.3 & 82.1 & 88.0   & 94.0   & 79.8 & 87.4 & 90.1\deltapos{2.5} \\
    \sys-Mimo-SFT (\(k=2\)) & 98.6 & 87.7 & 94.0   & 93.3 & 83.6 & 89.2 & 92.7 & 79.8 & 86.7 & 90.3\deltapos{2.7} \\
    \sys-Mimo-SFT (\(k=3\)) & 96.6 & 85.3 & 91.8 & 91.8 & 81.4 & 87.4 & 89.3 & 80.3 & 85.1 & 88.4\deltapos{0.8} \\
    \sys-Mimo-SFT (\(k=4\)) & 97.6 & 83.9 & 91.8 & 92.3 & 82.9 & 88.3 & 91.5 & 79.3 & 85.8 & 88.8\deltapos{1.2} \\ \midrule
    \rowcolor[HTML]{FFFBDE} 
    Mimo-VL-7B-RL        & 98.3 & 86.3 & 93.2 & 90.2 & 80.7 & 86.2 & 92.7 & 75.4 & 84.7 & 88.4        \\
    \sys-Mimo-RL (\(k=1\))  & 99.0   & 87.7 & 94.2 & 91.2 & 83.6 & 88.0   & 89.7 & 72.9 & 81.9 & 88.4\deltapos{0.0} \\
    \sys-Mimo-RL (\(k=2\))  & 97.9 & 86.7 & 93.2 & 90.7 & 80.7 & 86.5 & 91.0   & 75.4 & 83.8 & 88.2\deltaneg{0.2} \\
    \sys-Mimo-RL (\(k=3\))  & 99.0   & 87.2 & 94.0   & 91.8 & 80.7 & 87.1 & 91.0   & 76.8 & 84.4 & 88.9\deltapos{0.5} \\
    \sys-Mimo-RL (\(k=4\))  & 98.6 & 85.3 & 93   & 91.8 & 81.4 & 87.4 & 91.9 & 77.3 & 85.1 & 88.8\deltapos{0.4} \\ \bottomrule
    \end{tabular}%
    }
    \end{table}

%% file: tables/appendix-ss-pro-details.tex
\begin{table}[ht]
    \centering
    \caption{
      Performance on GUI grounding benchmark: ScreenSpot-Pro.
      Models are fine-tuned with 2K \(k\)-step UI Transition samples for each \(k \in \{1,2,3,4\}\), \textbf{without model-specific data filtering}.
    }
    \vspace{1em}
    \setlength{\tabcolsep}{3pt}
    \label{tab:appendix-ss-pro-details}
    \resizebox{\textwidth}{!}{%
    \begin{tabular}{@{}lccccccccccccc@{}}
    \toprule
     &
      \multicolumn{2}{c}{\textbf{CAD}} &
      \multicolumn{2}{c}{\textbf{Dev}} &
      \multicolumn{2}{c}{\textbf{Creative}} &
      \multicolumn{2}{c}{\textbf{Scientific}} &
      \multicolumn{2}{c}{\textbf{Office}} &
      \multicolumn{2}{c}{\textbf{OS}} &
       \\ \cmidrule(lr){2-3} \cmidrule(lr){4-5} \cmidrule(lr){6-7} \cmidrule(lr){8-9} \cmidrule(lr){10-11} \cmidrule(lr){12-13}
    
    \multirow{-2}{*}{\textbf{Model}} &
      Text &
      Icon &
      Text &
      Icon &
      Text &
      Icon &
      Text &
      Icon &
      Text &
      Icon &
      Text &
      Icon &
      \multirow{-2}{*}{\textbf{Avg.}} \\ \midrule
    \rowcolor[HTML]{CBDCEB} 
    Qwen2.5-VL-7B & 16.2 & 1.6  & 44.2 & 2.1  & 36.9 & 7.7  & 47.9 & 8.2  & 53.7 & 18.9 & 36.4 & 7.9  & 26.4        \\
    \sys-Qwen (\(k=1\))     & 16.2 & 4.7  & 52.6 & 9.0  & 27.3 & 7.0  & 52.1 & 5.5  & 49.7 & 17.0 & 38.3 & 13.5 & 26.8\deltapos{0.4} \\
    \sys-Qwen (\(k=2\))     & 16.8 & 3.1  & 52.6 & 10.3 & 29.3 & 8.4  & 52.1 & 4.5  & 50.8 & 17.0 & 35.5 & 13.5 & 27.2\deltapos{0.8} \\
    \sys-Qwen (\(k=3\))     & 15.7 & 3.1  & 52.6 & 9.7  & 30.3 & 7.0  & 50.7 & 5.5  & 48.0 & 15.1 & 35.5 & 12.4 & 26.5\deltapos{0.1} \\
    \sys-Qwen (\(k=4\))     & 17.3 & 3.1  & 51.9 & 9.7  & 30.3 & 7.0  & 54.2 & 5.5  & 49.7 & 17.0 & 33.6 & 12.4 & 27.1\deltapos{0.7} \\ \midrule
    \rowcolor[HTML]{E7F2EF} 
    InternVL3-8B         & 8.6  & 4.7  & 27.3 & 4.1  & 27.3 & 0.7  & 24.3 & 4.5  & 32.2 & 3.8  & 11.2 & 3.4  & 15.0        \\
    \sys-Intern (\(k=1\))   & 10.2 & 1.6  & 27.3 & 4.8  & 26.8 & 0.7  & 23.6 & 3.6  & 33.9 & 7.5  & 15.0 & 2.2  & 15.4\deltapos{0.4} \\
    \sys-Intern (\(k=2\))   & 7.6  & 3.1  & 27.3 & 3.4  & 28.3 & 0.7  & 21.5 & 4.5  & 33.3 & 7.5  & 12.1 & 2.2  & 14.9\deltaneg{0.1} \\
    \sys-Intern (\(k=3\))   & 9.6  & 3.1  & 23.4 & 5.5  & 26.3 & 0.7  & 20.1 & 1.8  & 29.9 & 3.8  & 11.2 & 1.1  & 13.7\deltaneg{1.3} \\
    \sys-Intern (\(k=4\))   & 9.1  & 4.7  & 24.0 & 4.1  & 28.8 & 0.7  & 21.5 & 3.6  & 31.1 & 3.8  & 13.1 & 1.1  & 14.5\deltaneg{0.5} \\ \midrule
    \rowcolor[HTML]{EFF5D2} 
    Mimo-VL-7B-SFT       & 47.2 & 23.4 & 48.7 & 9.0  & 48.0 & 13.3 & 70.8 & 27.3 & 64.4 & 39.6 & 36.4 & 15.7 & 39.8        \\
    \sys-Mimo-SFT (\(k=1\)) & 49.7 & 15.6 & 46.8 & 13.8 & 49.0 & 16.1 & 74.3 & 25.5 & 65.0 & 37.7 & 40.2 & 15.7 & 40.9\deltapos{1.1} \\
    \sys-Mimo-SFT (\(k=2\)) & 48.7 & 20.3 & 50.6 & 11.7 & 48.5 & 14.0 & 72.9 & 26.4 & 63.3 & 41.5 & 36.4 & 16.9 & 40.6\deltapos{0.8} \\
    \sys-Mimo-SFT (\(k=3\)) & 45.7 & 18.8 & 48.1 & 12.4 & 45.5 & 12.6 & 70.1 & 28.2 & 65.5 & 39.6 & 35.5 & 19.1 & 39.6\deltaneg{0.2} \\
    \sys-Mimo-SFT (\(k=4\)) & 45.7 & 18.8 & 51.3 & 12.4 & 46.5 & 12.6 & 71.5 & 28.2 & 62.7 & 41.5 & 37.4 & 16.9 & 39.9\deltapos{0.1} \\ \midrule
    \rowcolor[HTML]{FFFBDE} 
    Mimo-VL-7B-RL        & 48.2 & 14.1 & 46.8 & 11.0 & 47.0 & 14.0 & 71.5 & 27.3 & 66.7 & 39.6 & 39.3 & 19.1 & 40.2        \\
    \sys-Mimo-RL (\(k=1\))  & 48.7 & 15.6 & 46.1 & 13.8 & 51.0 & 13.3 & 72.2 & 30.0 & 66.7 & 43.4 & 42.1 & 21.3 & 41.7\deltapos{1.5} \\
    \sys-Mimo-RL (\(k=2\))  & 46.7 & 14.1 & 47.4 & 13.8 & 49.0 & 12.6 & 70.1 & 27.3 & 65.5 & 39.6 & 41.1 & 21.3 & 40.5\deltapos{0.3} \\
    \sys-Mimo-RL (\(k=3\))  & 49.7 & 12.5 & 47.4 & 11.0 & 47.0 & 11.2 & 70.8 & 26.4 & 65.5 & 45.3 & 35.5 & 20.2 & 39.9\deltaneg{0.3} \\
    \sys-Mimo-RL (\(k=4\))  & 46.7 & 14.1 & 46.1 & 13.8 & 47.0 & 14.7 & 69.4 & 27.3 & 64.4 & 41.5 & 36.4 & 21.3 & 39.8\deltaneg{0.4} \\ \bottomrule
    \end{tabular}%
    }
    \end{table}

%% file: tables/appendix-ss-pro-details-with-filter.tex
\begin{table}[ht]
    \centering
    \caption{
      Performance on GUI grounding benchmark: ScreenSpot-Pro.
      \textbf{Model-specific data filtering is applied} to InternVL3-8B, MimoVL-7B-SFT, and MimoVL-7B-RL.
      For each model, we select 2K \(k\)-step UI Transition samples for each \(k \in \{1,2,3,4\}\) from a pool of candidates.
    }
    \vspace{1em}
    \setlength{\tabcolsep}{3pt}
    \label{tab:appendix-ss-pro-details-with-filter}
    \resizebox{\textwidth}{!}{%
    \begin{tabular}{@{}lccccccccccccc@{}}
    \toprule
     &
      \multicolumn{2}{c}{\textbf{CAD}} &
      \multicolumn{2}{c}{\textbf{Dev}} &
      \multicolumn{2}{c}{\textbf{Creative}} &
      \multicolumn{2}{c}{\textbf{Scientific}} &
      \multicolumn{2}{c}{\textbf{Office}} &
      \multicolumn{2}{c}{\textbf{OS}} &
       \\ \cmidrule(lr){2-3} \cmidrule(lr){4-5} \cmidrule(lr){6-7} \cmidrule(lr){8-9} \cmidrule(lr){10-11} \cmidrule(lr){12-13}
    \multirow{-2}{*}{\textbf{Model}} &
      Text &
      Icon &
      Text &
      Icon &
      Text &
      Icon &
      Text &
      Icon &
      Text &
      Icon &
      Text &
      Icon &
      \multirow{-2}{*}{\textbf{Avg.}} \\ \midrule
    \rowcolor[HTML]{E7F2EF} 
    InternVL3-8B         & 8.6  & 4.7  & 27.3 & 4.1  & 27.3 & 0.7  & 24.3 & 4.5  & 32.2 & 3.8  & 11.2 & 3.4  & 15.0        \\
    \sys-Intern (\(k=1\))   & 11.2 & 3.1  & 26.6 & 5.5  & 28.8 & 0.7  & 20.1 & 4.5  & 33.9 & 7.5  & 13.1 & 1.1  & 15.4\deltapos{0.4} \\
    \sys-Intern (\(k=2\))   & 8.1   & 1.6  & 26.0 & 4.1  & 26.3 & 0.7  & 25.7 & 1.8  & 31.1 & 5.7  & 14.0 & 1.1  & 14.5\deltaneg{0.5} \\
    \sys-Intern (\(k=3\))   & 8.6  & 1.6  & 27.3 & 4.1  & 25.3 & 0.7  & 21.5 & 1.8  & 33.3 & 5.7  & 14.0 & 1.1  & 14.4\deltaneg{0.6} \\
    \sys-Intern (\(k=4\))   & 8.1  & 1.6  & 27.9 & 4.8  & 26.8 & 0.0  & 22.9 & 2.7  & 32.8 & 3.8  & 10.3 & 2.2  & 14.5\deltaneg{0.5} \\ \midrule
    \rowcolor[HTML]{EFF5D2} 
    Mimo-VL-7B-SFT       & 47.2 & 23.4 & 48.7 & 9.0  & 48.0 & 13.3 & 70.8 & 27.3 & 64.4 & 39.6 & 36.4 & 15.7 & 39.8        \\
    \sys-Mimo-SFT (\(k=1\)) & 49.2 & 14.1 & 50.0 & 9.0 & 48.0 & 11.9 & 73.6 & 29.1 & 54.3 & 43.4 & 38.3 & 15.7 & 40.7\deltapos{0.9} \\
    \sys-Mimo-SFT (\(k=2\)) & 46.2 & 17.2 & 46.8 & 11.0   & 47.5 & 10.5 & 74.3 & 30.9 & 65.5 & 43.4 & 35.5 & 16.9 & 40.0\deltapos{0.2} \\
    \sys-Mimo-SFT (\(k=3\)) & 50.8 & 15.6 & 47.4 & 13.8 & 47.0   & 12.6 & 69.4 & 24.5 & 65.0   & 41.5 & 38.3 & 19.1 & 40.2\deltapos{0.4} \\
    \sys-Mimo-SFT (\(k=4\)) & 44.2 & 18.8 & 53.9 & 11.0   & 43.9 & 11.2 & 72.2 & 26.4 & 62.1 & 47.2 & 40.2 & 12.4 & 39.4\deltaneg{0.4} \\ \midrule
    \rowcolor[HTML]{FFFBDE} 
    Mimo-VL-7B-RL        & 48.2 & 14.1 & 46.8 & 11.0 & 47.0 & 14.0 & 71.5 & 27.3 & 66.7 & 39.6 & 39.3 & 19.1 & 40.2        \\
    \sys-Mimo-RL (\(k=1\))  & 48.7 & 14.1 & 51.9 & 13.1 & 50.0   & 15.4 & 70.8 & 28.2 & 67.2 & 39.6 & 37.4 & 21.3 & 41.6\deltapos{1.4} \\
    \sys-Mimo-RL (\(k=2\))  & 47.2 & 18.8 & 47.4 & 14.5 & 49.0   & 15.4 & 72.2 & 29.1 & 66.1 & 43.4 & 39.3 & 19.1 & 41.3\deltapos{0.9} \\
    \sys-Mimo-RL (\(k=3\))  & 45.7 & 17.2 & 46.1 & 11.7 & 49.5 & 14.0   & 70.8 & 29.1 & 66.1 & 39.6 & 37.4 & 21.3 & 40.4\deltapos{0.2} \\
    \sys-Mimo-RL (\(k=4\))  & 47.7 & 18.8 & 48.1 & 13.8 & 52.5 & 12.6 & 72.2 & 28.2 & 67.2 & 45.3 & 40.2 & 20.2 & 41.8\deltapos{1.6} \\ \bottomrule
    \end{tabular}%
    }
    \end{table}

%% file: main-iclr.bbl
\begin{thebibliography}{40}
\providecommand{\natexlab}[1]{#1}
\providecommand{\url}[1]{\texttt{#1}}
\expandafter\ifx\csname urlstyle\endcsname\relax
  \providecommand{\doi}[1]{doi: #1}\else
  \providecommand{\doi}{doi: \begingroup \urlstyle{rm}\Url}\fi

\bibitem[Bai et~al.(2025)Bai, Chen, Liu, Wang, Ge, Song, Dang, Wang, Wang, Tang, et~al.]{bai2025qwen25vltechnicalreport}
Shuai Bai, Keqin Chen, Xuejing Liu, Jialin Wang, Wenbin Ge, Sibo Song, Kai Dang, Peng Wang, Shijie Wang, Jun Tang, et~al.
\newblock Qwen2. 5-vl technical report.
\newblock \emph{arXiv preprint arXiv:2502.13923}, 2025.

\bibitem[Brandfonbrener et~al.(2023)Brandfonbrener, Nachum, and Bruna]{brandfonbrener2023inverse}
David Brandfonbrener, Ofir Nachum, and Joan Bruna.
\newblock Inverse dynamics pretraining learns good representations for multitask imitation.
\newblock In \emph{NeurIPS}, 2023.

\bibitem[Chen et~al.(2024)Chen, Wu, Wang, Su, Chen, Xing, Zhong, Zhang, Zhu, Lu, et~al.]{chen2024internvlscalingvisionfoundation}
Zhe Chen, Jiannan Wu, Wenhai Wang, Weijie Su, Guo Chen, Sen Xing, Muyan Zhong, Qinglong Zhang, Xizhou Zhu, Lewei Lu, et~al.
\newblock Internvl: Scaling up vision foundation models and aligning for generic visual-linguistic tasks.
\newblock In \emph{Proceedings of the IEEE/CVF conference on computer vision and pattern recognition}, pp.\  24185--24198, 2024.

\bibitem[Cheng et~al.(2024)Cheng, Sun, Chu, Xu, Li, Zhang, and Wu]{cheng2024seeclickharnessingguigrounding}
Kanzhi Cheng, Qiushi Sun, Yougang Chu, Fangzhi Xu, Yantao Li, Jianbing Zhang, and Zhiyong Wu.
\newblock Seeclick: Harnessing {GUI} grounding for advanced visual {GUI} agents.
\newblock In \emph{{ACL} {(1)}}, pp.\  9313--9332. Association for Computational Linguistics, 2024.

\bibitem[Dai et~al.(2025)Dai, Jiang, Cao, Li, Yang, Tan, Li, and Qiu]{dai2025advancingmobileguiagents}
Gaole Dai, Shiqi Jiang, Ting Cao, Yuanchun Li, Yuqing Yang, Rui Tan, Mo~Li, and Lili Qiu.
\newblock Advancing mobile gui agents: A verifier-driven approach to practical deployment.
\newblock \emph{arXiv preprint arXiv:2503.15937}, 2025.

\bibitem[Deka et~al.(2017)Deka, Huang, Franzen, Hibschman, Afergan, Li, Nichols, and Kumar]{Deka:2017:Rico}
Biplab Deka, Zifeng Huang, Chad Franzen, Joshua Hibschman, Daniel Afergan, Yang Li, Jeffrey Nichols, and Ranjitha Kumar.
\newblock Rico: {A} mobile app dataset for building data-driven design applications.
\newblock In \emph{{UIST}}, pp.\  845--854. {ACM}, 2017.

\bibitem[Gao et~al.(2024)Gao, Zhang, Wang, Wang, Li, and Xu]{gao2024mobileviewslargescalemobilegui}
Longxi Gao, Li~Zhang, Shihe Wang, Shangguang Wang, Yuanchun Li, and Mengwei Xu.
\newblock Mobileviews: A large-scale mobile gui dataset.
\newblock \emph{arXiv preprint arXiv:2409.14337}, 2024.

\bibitem[Gou et~al.(2024)Gou, Wang, Zheng, Xie, Chang, Shu, Sun, and Su]{gou2025navigatingdigitalworldhumans}
Boyu Gou, Ruohan Wang, Boyuan Zheng, Yanan Xie, Cheng Chang, Yiheng Shu, Huan Sun, and Yu~Su.
\newblock Navigating the digital world as humans do: Universal visual grounding for gui agents.
\newblock \emph{arXiv preprint arXiv:2410.05243}, 2024.

\bibitem[Gu et~al.(2025)Gu, Zeng, Xu, Zhou, Shen, Liu, Zhou, Meng, Xia, Chen, et~al.]{gu2025uivenus}
Zhangxuan Gu, Zhengwen Zeng, Zhenyu Xu, Xingran Zhou, Shuheng Shen, Yunfei Liu, Beitong Zhou, Changhua Meng, Tianyu Xia, Weizhi Chen, et~al.
\newblock Ui-venus technical report: Building high-performance ui agents with rft.
\newblock \emph{arXiv preprint arXiv:2508.10833}, 2025.

\bibitem[Guo et~al.(2025)Guo, Yang, Zhang, Song, Zhang, Xu, Zhu, Ma, Wang, Bi, et~al.]{deepseekai2025deepseekr1incentivizingreasoningcapability}
Daya Guo, Dejian Yang, Haowei Zhang, Junxiao Song, Ruoyu Zhang, Runxin Xu, Qihao Zhu, Shirong Ma, Peiyi Wang, Xiao Bi, et~al.
\newblock Deepseek-r1: Incentivizing reasoning capability in llms via reinforcement learning.
\newblock \emph{arXiv preprint arXiv:2501.12948}, 2025.

\bibitem[Hong et~al.(2024)Hong, Wang, Lv, Xu, Yu, Ji, Wang, Wang, Dong, Ding, and Tang]{hong2024cogagentvisuallanguagemodel}
Wenyi Hong, Weihan Wang, Qingsong Lv, Jiazheng Xu, Wenmeng Yu, Junhui Ji, Yan Wang, Zihan Wang, Yuxiao Dong, Ming Ding, and Jie Tang.
\newblock Cogagent: {A} visual language model for {GUI} agents.
\newblock In \emph{{CVPR}}, pp.\  14281--14290. {IEEE}, 2024.

\bibitem[Lambert et~al.(2024)Lambert, Morrison, Pyatkin, Huang, Ivison, Brahman, Miranda, Liu, Dziri, Lyu, et~al.]{lambert2025tulu3pushingfrontiers}
Nathan Lambert, Jacob Morrison, Valentina Pyatkin, Shengyi Huang, Hamish Ivison, Faeze Brahman, Lester James~V Miranda, Alisa Liu, Nouha Dziri, Shane Lyu, et~al.
\newblock T$\backslash$" ulu 3: Pushing frontiers in open language model post-training.
\newblock \emph{arXiv preprint arXiv:2411.15124}, 2024.

\bibitem[Li et~al.(2025{\natexlab{a}})Li, Meng, Lin, Luo, Tian, Ma, Huang, and Chua]{li2025screenspotproguigroundingprofessional}
Kaixin Li, Ziyang Meng, Hongzhan Lin, Ziyang Luo, Yuchen Tian, Jing Ma, Zhiyong Huang, and Tat-Seng Chua.
\newblock Screenspot-pro: Gui grounding for professional high-resolution computer use.
\newblock \emph{arXiv preprint arXiv:2504.07981}, 2025{\natexlab{a}}.

\bibitem[Li et~al.(2025{\natexlab{b}})Li, Ziyang, Lin, Luo, Tian, Ma, Huang, and Chua]{li2025screenspotpro}
Kaixin Li, Meng Ziyang, Hongzhan Lin, Ziyang Luo, Yuchen Tian, Jing Ma, Zhiyong Huang, and Tat-Seng Chua.
\newblock Screenspot-pro: {GUI} grounding for professional high-resolution computer use.
\newblock In \emph{Workshop on Reasoning and Planning for Large Language Models}, 2025{\natexlab{b}}.
\newblock URL \url{https://openreview.net/forum?id=XaKNDIAHas}.

\bibitem[Li et~al.(2024)Li, Bishop, Li, Rawles, Campbell{-}Ajala, Tyamagundlu, and Riva]{li2024effectsdatascaleui}
Wei Li, William~E. Bishop, Alice Li, Christopher Rawles, Folawiyo Campbell{-}Ajala, Divya Tyamagundlu, and Oriana Riva.
\newblock On the effects of data scale on {UI} control agents.
\newblock In \emph{NeurIPS}, 2024.

\bibitem[Lin et~al.(2024)Lin, Li, Gao, Yang, Wu, Bai, Lei, Wang, and Shou]{lin2024showui}
Kevin~Qinghong Lin, Linjie Li, Difei Gao, Zhengyuan Yang, Shiwei Wu, Zechen Bai, Weixian Lei, Lijuan Wang, and Mike~Zheng Shou.
\newblock Showui: One vision-language-action model for gui visual agent.
\newblock \emph{arXiv preprint arXiv:2411.17465}, 2024.

\bibitem[Liu et~al.(2025)Liu, Li, Xie, Hu, Han, Zhang, Yang, and Wu]{liu2025infiguir1advancingmultimodalgui}
Yuhang Liu, Pengxiang Li, Congkai Xie, Xavier Hu, Xiaotian Han, Shengyu Zhang, Hongxia Yang, and Fei Wu.
\newblock Infigui-r1: Advancing multimodal gui agents from reactive actors to deliberative reasoners.
\newblock \emph{arXiv preprint arXiv:2504.14239}, 2025.

\bibitem[Lu et~al.(2024)Lu, Shao, Liu, Meng, Li, Chen, Huang, Zhang, Qiao, and Luo]{lu2024guiodyssey}
Quanfeng Lu, Wenqi Shao, Zitao Liu, Fanqing Meng, Boxuan Li, Botong Chen, Siyuan Huang, Kaipeng Zhang, Yu~Qiao, and Ping Luo.
\newblock Gui odyssey: A comprehensive dataset for cross-app gui navigation on mobile devices.
\newblock \emph{arXiv preprint arXiv:2406.08451}, 2024.

\bibitem[Lu et~al.(2025)Lu, Chai, Guo, Yin, Liu, Wang, Xiao, Ren, Xiong, and Li]{lu2025uir1enhancingactionprediction}
Zhengxi Lu, Yuxiang Chai, Yaxuan Guo, Xi~Yin, Liang Liu, Hao Wang, Han Xiao, Shuai Ren, Guanjing Xiong, and Hongsheng Li.
\newblock Ui-r1: Enhancing action prediction of gui agents by reinforcement learning.
\newblock \emph{arXiv preprint arXiv:2503.21620}, 2025.

\bibitem[OpenAI(2024)]{hurst2024gpt}
OpenAI.
\newblock Gpt-4o system card.
\newblock \emph{arXiv preprint arXiv:2410.21276}, 2024.

\bibitem[Qin et~al.(2025)Qin, Ye, Fang, Wang, Liang, Tian, Zhang, Li, Li, Huang, et~al.]{qin2025uitarspioneeringautomatedgui}
Yujia Qin, Yining Ye, Junjie Fang, Haoming Wang, Shihao Liang, Shizuo Tian, Junda Zhang, Jiahao Li, Yunxin Li, Shijue Huang, et~al.
\newblock Ui-tars: Pioneering automated gui interaction with native agents.
\newblock \emph{arXiv preprint arXiv:2501.12326}, 2025.

\bibitem[Rawles et~al.(2023)Rawles, Li, Rodriguez, Riva, and Lillicrap]{rawles2023androidwildlargescaledataset}
Christopher Rawles, Alice Li, Daniel Rodriguez, Oriana Riva, and Timothy Lillicrap.
\newblock Androidinthewild: A large-scale dataset for android device control.
\newblock \emph{Advances in Neural Information Processing Systems}, 36:\penalty0 59708--59728, 2023.

\bibitem[Rawles et~al.(2024)Rawles, Clinckemaillie, Chang, Waltz, Lau, Fair, Li, Bishop, Li, Campbell-Ajala, et~al.]{rawles2024androidworld}
Christopher Rawles, Sarah Clinckemaillie, Yifan Chang, Jonathan Waltz, Gabrielle Lau, Marybeth Fair, Alice Li, William Bishop, Wei Li, Folawiyo Campbell-Ajala, et~al.
\newblock Androidworld: A dynamic benchmarking environment for autonomous agents.
\newblock \emph{arXiv preprint arXiv:2405.14573}, 2024.

\bibitem[Schulman et~al.(2017)Schulman, Wolski, Dhariwal, Radford, and Klimov]{schulman2017proximalpolicyoptimizationalgorithms}
John Schulman, Filip Wolski, Prafulla Dhariwal, Alec Radford, and Oleg Klimov.
\newblock Proximal policy optimization algorithms.
\newblock \emph{arXiv preprint arXiv:1707.06347}, 2017.

\bibitem[Shao et~al.(2024)Shao, Wang, Zhu, Xu, Song, Bi, Zhang, Zhang, Li, Wu, et~al.]{shao2024deepseekmathpushinglimitsmathematical}
Zhihong Shao, Peiyi Wang, Qihao Zhu, Runxin Xu, Junxiao Song, Xiao Bi, Haowei Zhang, Mingchuan Zhang, YK~Li, Y~Wu, et~al.
\newblock Deepseekmath: Pushing the limits of mathematical reasoning in open language models.
\newblock \emph{arXiv preprint arXiv:2402.03300}, 2024.

\bibitem[Shen et~al.(2025)Shen, Liu, Li, Fang, Ma, Liao, Shen, Zhang, Zhao, Zhang, et~al.]{shen2025vlmr1}
Haozhan Shen, Peng Liu, Jingcheng Li, Chunxin Fang, Yibo Ma, Jiajia Liao, Qiaoli Shen, Zilun Zhang, Kangjia Zhao, Qianqian Zhang, et~al.
\newblock Vlm-r1: A stable and generalizable r1-style large vision-language model.
\newblock \emph{arXiv preprint arXiv:2504.07615}, 2025.

\bibitem[Tang et~al.(2025)Tang, Xia, Wu, Hu, Chen, Chen, Xu, Wu, Lu, Ma, et~al.]{tang2025lpo}
Jiaqi Tang, Yu~Xia, Yi-Feng Wu, Yuwei Hu, Yuhui Chen, Qing-Guo Chen, Xiaogang Xu, Xiangyu Wu, Hao Lu, Yanqing Ma, et~al.
\newblock Lpo: Towards accurate gui agent interaction via location preference optimization.
\newblock \emph{arXiv preprint arXiv:2506.09373}, 2025.

\bibitem[Tian et~al.(2024)Tian, Yang, Zeng, Wang, Lin, Dong, and Pang]{tian2024predictive}
Yang Tian, Sizhe Yang, Jia Zeng, Ping Wang, Dahua Lin, Hao Dong, and Jiangmiao Pang.
\newblock Predictive inverse dynamics models are scalable learners for robotic manipulation.
\newblock \emph{arXiv preprint arXiv:2412.15109}, 2024.

\bibitem[Wang et~al.(2024)Wang, Lv, Yu, Hong, Qi, Wang, Ji, Yang, Zhao, Song, Xu, Chen, Xu, Li, Dong, Ding, and Tang]{wang2023cogvlm}
Weihan Wang, Qingsong Lv, Wenmeng Yu, Wenyi Hong, Ji~Qi, Yan Wang, Junhui Ji, Zhuoyi Yang, Lei Zhao, Xixuan Song, Jiazheng Xu, Keqin Chen, Bin Xu, Juanzi Li, Yuxiao Dong, Ming Ding, and Jie Tang.
\newblock Cogvlm: Visual expert for pretrained language models.
\newblock In \emph{NeurIPS}, 2024.

\bibitem[Wen et~al.(2024)Wen, Li, Liu, Zhao, Yu, Li, Jiang, Liu, Zhang, and Liu]{wen2024autodroidllmpoweredtaskautomation}
Hao Wen, Yuanchun Li, Guohong Liu, Shanhui Zhao, Tao Yu, Toby~Jia{-}Jun Li, Shiqi Jiang, Yunhao Liu, Yaqin Zhang, and Yunxin Liu.
\newblock Autodroid: Llm-powered task automation in android.
\newblock In \emph{MobiCom}, pp.\  543--557. {ACM}, 2024.

\bibitem[Wu et~al.(2024{\natexlab{a}})Wu, Xu, Liu, Tan, Liu, Li, Luan, Wang, and Shang]{wu2024mobilevlmvisionlanguagemodelbetter}
Qinzhuo Wu, Weikai Xu, Wei Liu, Tao Tan, Jianfeng Liu, Ang Li, Jian Luan, Bin Wang, and Shuo Shang.
\newblock Mobilevlm: {A} vision-language model for better intra- and inter-ui understanding.
\newblock In \emph{{EMNLP} (Findings)}, pp.\  10231--10251. Association for Computational Linguistics, 2024{\natexlab{a}}.

\bibitem[Wu et~al.(2024{\natexlab{b}})Wu, Wu, Xu, Wang, Sun, Jia, Cheng, Ding, Chen, Liang, et~al.]{wu2024osatlasfoundationactionmodel}
Zhiyong Wu, Zhenyu Wu, Fangzhi Xu, Yian Wang, Qiushi Sun, Chengyou Jia, Kanzhi Cheng, Zichen Ding, Liheng Chen, Paul~Pu Liang, et~al.
\newblock Os-atlas: A foundation action model for generalist gui agents.
\newblock \emph{arXiv preprint arXiv:2410.23218}, 2024{\natexlab{b}}.

\bibitem[Xia \& Luo(2025)Xia and Luo]{luo2025guir1generalistr1style}
Xiaobo Xia and Run Luo.
\newblock Gui-r1: A generalist r1-style vision-language action model for gui agents.
\newblock \emph{arXiv preprint arXiv:2504.10458}, 2025.

\bibitem[Xiaomi(2025{\natexlab{a}})]{xiaomi-guievalkit2025}
Xiaomi.
\newblock Guievalkit: Open-source evaluation toolkit for gui agents.
\newblock \url{https://github.com/xiaomi-research/guievalkit}, 2025{\natexlab{a}}.
\newblock Accessed: 2025-09-25.

\bibitem[Xiaomi(2025{\natexlab{b}})]{coreteam2025mimovltechnicalreport}
LLM-Core-Team Xiaomi.
\newblock Mimo-vl technical report.
\newblock \emph{arXiv preprint arXiv:2506.03569}, 2025{\natexlab{b}}.

\bibitem[Xu et~al.(2024)Xu, Wang, Wang, Lu, Xie, Saha, Sahoo, Yu, and Xiong]{xu2024aguvis}
Yiheng Xu, Zekun Wang, Junli Wang, Dunjie Lu, Tianbao Xie, Amrita Saha, Doyen Sahoo, Tao Yu, and Caiming Xiong.
\newblock Aguvis: Unified pure vision agents for autonomous gui interaction.
\newblock \emph{arXiv preprint arXiv:2412.04454}, 2024.

\bibitem[Yang et~al.(2024)Yang, Wang, Li, Luo, Chen, Huang, and Li]{yang2024ariauivisualgroundinggui}
Yuhao Yang, Yue Wang, Dongxu Li, Ziyang Luo, Bei Chen, Chao Huang, and Junnan Li.
\newblock Aria-ui: Visual grounding for gui instructions.
\newblock \emph{arXiv preprint arXiv:2412.16256}, 2024.

\bibitem[Zapolsky \& Drumwright(2017)Zapolsky and Drumwright]{zapolsky2016inversedynamicsrigidcontact}
Samuel Zapolsky and Evan~M. Drumwright.
\newblock Inverse dynamics with rigid contact and friction.
\newblock \emph{Auton. Robots}, 41\penalty0 (4):\penalty0 831--863, 2017.

\bibitem[Zhang et~al.(2025{\natexlab{a}})Zhang, Gao, and Xu]{zhang2025doeschainofthoughtreasoninghelp}
Li~Zhang, Longxi Gao, and Mengwei Xu.
\newblock Does chain-of-thought reasoning help mobile gui agent? an empirical study.
\newblock \emph{arXiv preprint arXiv:2503.16788}, 2025{\natexlab{a}}.

\bibitem[Zhang et~al.(2025{\natexlab{b}})Zhang, Lu, Fu, Huo, Yang, Wu, Si, Cong, Chen, Lin, Xie, Zhou, Xu, Zhang, Su, Zhai, Liu, Mei, Xu, Tian, Wang, Chen, Yao, Liu, and Sun]{zhang2025agentcpmgui}
Zhong Zhang, Yaxi Lu, Yikun Fu, Yupeng Huo, Shenzhi Yang, Yesai Wu, Han Si, Xin Cong, Haotian Chen, Yankai Lin, Jie Xie, Wei Zhou, Wang Xu, Yuanheng Zhang, Zhou Su, Zhongwu Zhai, Xiaoming Liu, Yudong Mei, Jianming Xu, Hongyan Tian, Chongyi Wang, Chi Chen, Yuan Yao, Zhiyuan Liu, and Maosong Sun.
\newblock Agent{CPM}-{GUI}: Building mobile-use agents with reinforcement fine-tuning.
\newblock \emph{arXiv preprint arXiv:2506.01391}, 2025{\natexlab{b}}.

\end{thebibliography}
